\documentclass[10pt,twocolumn,letterpaper]{article}

\usepackage{configs/cvpr}              

\usepackage{float}
\usepackage[accsupp]{axessibility}
\usepackage{graphicx}
\usepackage{amsmath}
\usepackage{amssymb}
\usepackage{multirow}
\usepackage{booktabs}
\usepackage{cite}
\usepackage{amsmath,amssymb,amsfonts}
\usepackage{algorithmic}
\usepackage{tikz}
\usepackage{url}
\usepackage{amsmath}
\usepackage{makecell}
\usepackage{tabularx}
\usepackage{pgfplots}
\usepackage{graphicx}
\usepackage{multirow}
\usepackage{adjustbox}
\usepackage{subcaption}
\usepackage{caption}
\usepackage{float}
\usepackage{textcomp}
\usepackage{xcolor}
\usepackage{pgfplots}       

\usepackage[pagebackref,breaklinks,colorlinks]{hyperref}

\usepackage[capitalize]{cleveref}
\crefname{section}{Sec.}{Secs.}
\Crefname{section}{Section}{Sections}
\Crefname{table}{Table}{Tables}
\crefname{table}{Tab.}{Tabs.}

\begin{document}

\title{NeutrEx: A 3D Quality Component Measure on Facial Expression Neutrality}

\author{Marcel Grimmer\\
NTNU\thanks{Norwegian University of Science and Technology}\\
Gj{\o}vik, Norway\\
{\tt\small marceg@ntnu.com}
\and 
Christian Rathgeb\\
h\_da\thanks{Hochschule Darmstadt}\\
Darmstadt, Germany\\
{\tt\small christian.rathgeb@h-da.de}
\and 
Raymond Veldhuis\\
UT\thanks{University of Twente}  \qquad NTNU\footnotemark[1]\\
Twente, Netherlands \qquad Gj{\o}vik, Norway\\
{\tt\small r.n.j.veldhuis@utwente.nl}
\and 
Christoph Busch\\
NTNU\footnotemark[1]  \qquad h\_da\footnotemark[2]\\
Gj{\o}vik, Norway \qquad Darmstadt, Germany\\
{\tt\small christoph.busch@ntnu.no}
}

\maketitle

\begin{abstract}

Accurate face recognition systems are increasingly important in sensitive applications like border control or migration management. Therefore, it becomes crucial to quantify the quality of facial images to ensure that low-quality images are not affecting recognition accuracy. In this context, the current draft of \textit{ISO/IEC 29794-5} introduces the concept of component quality to estimate how single factors of variation affect recognition outcomes. In this study, we propose a quality measure (NeutrEx) based on the accumulated distances of a 3D face reconstruction to a neutral expression anchor. Our evaluations demonstrate the superiority of our proposed method compared to baseline approaches obtained by training Support Vector Machines on face embeddings extracted from a pre-trained Convolutional Neural Network for facial expression classification. Furthermore, we highlight the explainable nature of our NeutrEx measures by computing per-vertex distances to unveil the most impactful face regions and allow operators to give actionable feedback to subjects\footnote{Code and pre-trained models available under \url{https://github.com/datasciencegrimmer/NeutrEx-quality-component}}.
\end{abstract}
\section{Introduction}
Face recognition (FR) systems are employed in various applications, such as border control~\cite{EU-Regulation-EES-InternalDocument-2017}, forensics~\cite{EU-Regulation-SIS-1862-2018}, and migration management~\cite{EU-Regulation-2008-767-on-VIS-090706}. To ensure reliable and convenient authentication, the recognition accuracy must comply with high standards~\cite{FRONTEX-BorderControl-BestPractices-InternalDocument-2015}.

The dependency between \textit{face image quality} and \textit{biometric recognition performance} is widely studied~\cite{schlett2022face} to strengthen the understanding of the operational boundaries in which FR systems can verify mated samples accurately and support interoperability across systems~\cite{EU-Regulation-2019-817-on-Interoperability-Framework-190520}. Motivated by these endeavours, the current working draft of \textit{ISO/IEC 29794-5}~\cite{ISO-IEC-29794-5-WD6-FaceQuality-230119} pursues to standardize face image quality.

\begin{figure}
\centering
  \begin{subfigure}[b]{0.175\linewidth}
    \includegraphics[width=\linewidth]{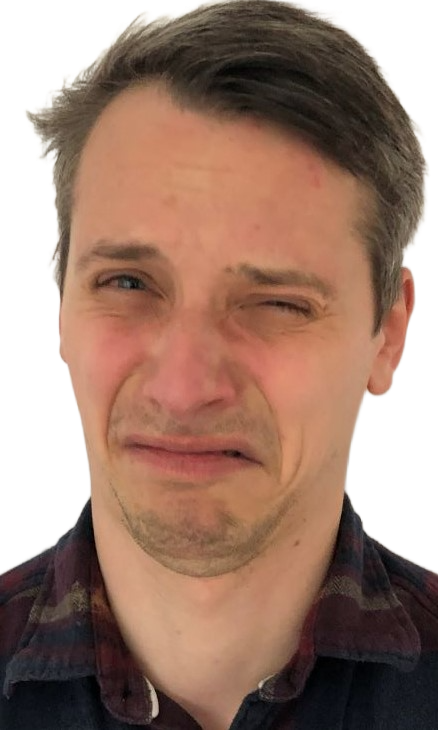}
    \caption{Input}
    \label{fig:synth-face-example}
  \end{subfigure}
  \begin{subfigure}[b]{0.2\linewidth}
    \includegraphics[width=\linewidth]{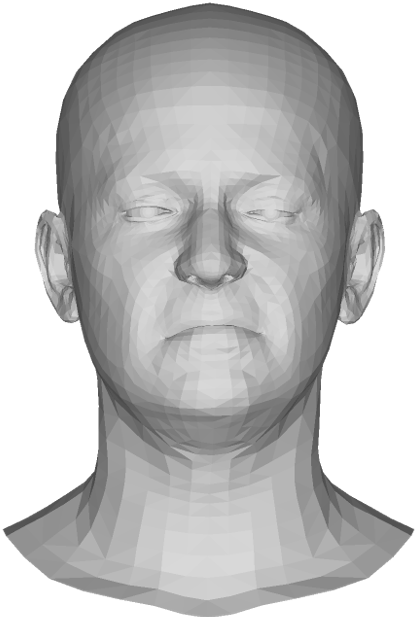}
    \caption{3D Mesh}
    \label{fig:synth-face-example}
  \end{subfigure}
    \begin{subfigure}[b]{0.2\linewidth}
    \includegraphics[width=\linewidth]{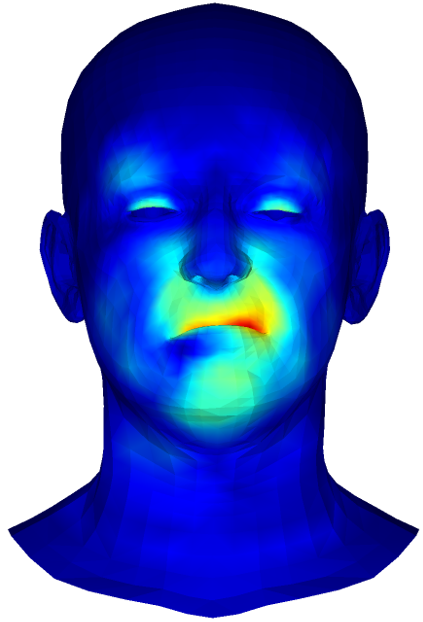}
    \caption{Residuals}
    \label{fig:synth-face-example}
  \end{subfigure}
    \begin{subfigure}[b]{0.2\linewidth}
    \includegraphics[width=\linewidth]{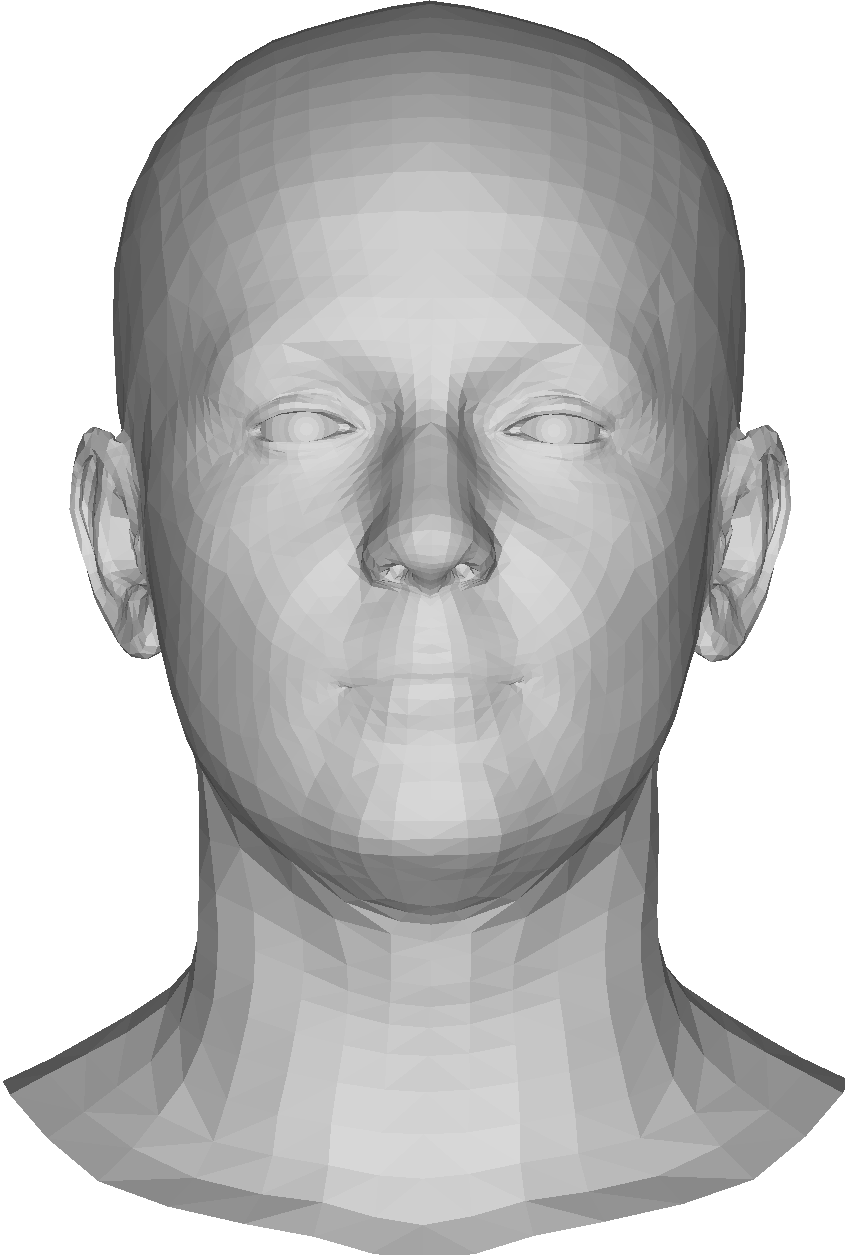}
    \caption{Anchor}
    \label{fig:real-face-example}
  \end{subfigure}

  \begin{subfigure}[b]{0.175\linewidth}
    \includegraphics[width=\linewidth]{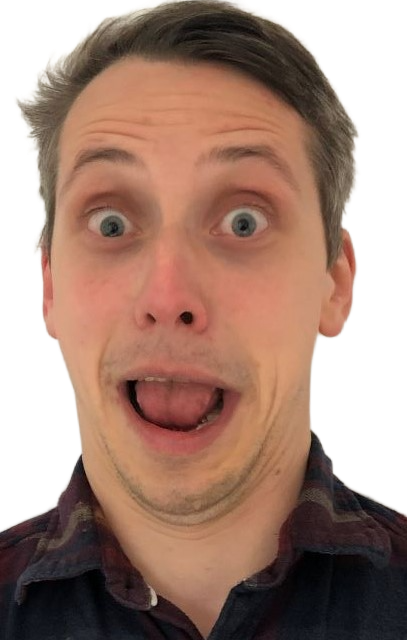}
    \caption{Input}
    \label{fig:real-face-example}
  \end{subfigure}
  \begin{subfigure}[b]{0.2\linewidth}
    \includegraphics[width=\linewidth]{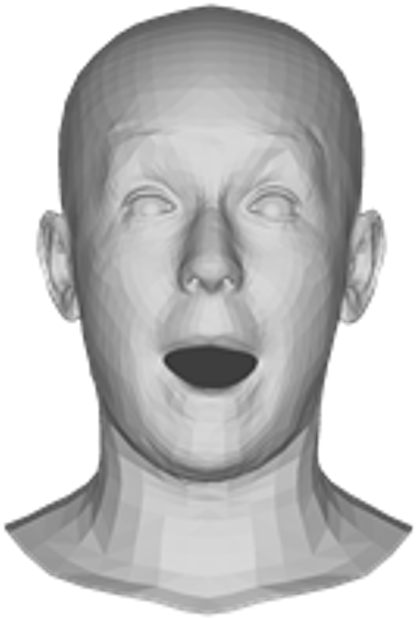}
    \caption{3D Mesh}
    \label{fig:real-face-example}
  \end{subfigure}
  \begin{subfigure}[b]{0.2\linewidth}
    \includegraphics[width=\linewidth]{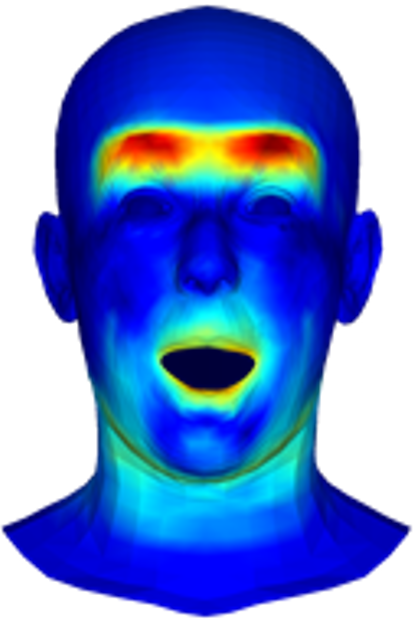}
    \caption{Residuals}
    \label{fig:real-face-example}
  \end{subfigure}
  \begin{subfigure}[b]{0.2\linewidth}
    \includegraphics[width=\linewidth]{images/neutral-reference.png}
    \caption{Anchor}
    \label{fig:real-face-example}
  \end{subfigure}
  \caption{Residual maps visualizing per-vertex Euclidean distances from the neutral anchor on two example 3D face models. \label{fig:intro-residual-examples}}
\end{figure}

\subsection{Face Image Quality Assessment}

In this work, we refer to biometric quality as the standardized term of \textit{utility}~\cite{ISO-IEC-29794-1-QualityFramework-2023} that reflects the predicted positive or negative contribution of an individual sample to the overall performance of a biometric system. Fundamentally, ISO/IEC 29794-5~\cite{ISO-IEC-29794-5-WD6-FaceQuality-230119} further divides biometric quality into two concepts: \textit{unified} and \textit{component quality}.
\textit{Unified quality scores} assess the overall face image quality and predict the recognition outcome, taking into account all factors of variation and their interrelations. In contrast, \textit{quality components} quantify how each individual \textit{capture-} or \textit{subject-related} measure affects the recognition performance. In practice, face image quality assessment (FIQA) algorithms are particularly used to 1) filter out low-quality samples automatically (unified quality) and 2) provide subjects with actionable feedback (component quality). Hence, unified and component quality can be treated as complementary to each other to improve the system's efficiency and explainability.         

\subsection{Quality Component: Expression Neutrality}
 
As shown in the literature~\cite{pena2021facial}\cite{damer2018crazyfaces}~\cite{merkle2022state}, extreme facial expressions significantly contribute to the number of false non-matches caused by the increased intra-identity variation. Moreover, extreme facial expressions were successfully used to prevent individuals from being recognized~\cite{damer2018crazyfaces}, potentially enabling them to intentionally deceive their identity.

To address this issue, the current committee draft of \textit{ISO/IEC 29794-5} defines \textit{facial expression neutrality} as a subject-related quality component. It is assumed that ICAO-compliant face images with neutral expressions represent the best-possible image quality that enables the extraction of identity features best suitable for recognition. According to the biometric sample quality framework standard~\cite{ISO-IEC-29794-1-QualityFramework-2023}, all component quality values shall be expressed in the range of $[0, 100]$, where $0$ and $100$ represent the lower and upper quality boundaries, respectively. Finding a mapping function from facial images to their respective expression neutrality measures is non-trivial, as expression neutrality itself naturally underlies variation and depends on factors such as an individual's head shape or ethnicity.   

\subsection{Contribution \& Paper Structure}

This work contributes to the development of a facial expression neutrality measure that is compliant with ISO/IEC 29794-5 by proposing a novel method (\textit{NeutrEx}) based on \textit{3D Morphable Face Models} (3DMMs)~\cite{Li-FLAME-ToG-2017} and recent advancements in \textit{Monocular 3D Face Reconstruction}~\cite{Feng-DECA-ToG-2021, Danvevcek-EMOCA-CVPR-2022}. In addition, we introduce two baseline approaches to validate the soundness of our NeutrEx measures.

Specifically, we combine the pre-trained \textit{coarse shape encoder} of Feng et al.~\cite{Feng-DECA-ToG-2021} (\textit{DECA}) with the \textit{expression encoder} of Danvevcek et al.~\cite{Danvevcek-EMOCA-CVPR-2022} (\textit{EMOCA}) to translate 2D face images into the parameter space of a 3DMM (\textit{FLAME}~\cite{Li-FLAME-ToG-2017}). Once encoded, we use the FLAME decoder to reconstruct the 3D face representation of each face comprised of 5,023 vertices.

To measure the deviation from expression neutrality, we derive a neutral expression anchor by computing the average 3D face mesh upon all neutral samples in the training datasets, namely \textit{Multi-PIE}~\cite{Gross-MultiPIE-IVC-2010} and \textit{FEAFA+}~\cite{GAN-FEAFA+-ICDIP-2022}. By leveraging the full vertex correspondence FLAME, we measure the average \textit{per-vertex} distance between a reconstructed 3D face model to the neutral anchor (see Figure~\ref{fig:intro-residual-examples}). To disentangle our NeutrEx measures from the effects of individual head poses and shapes, we normalize each 3D face model by rotating it to a frontal view and transforming it into a generic head shape.

In summary, the contributions of this work can be described as follows:

\begin{itemize}
    \item We propose NeutrEx: A 3D-based approach to measure the distance of a facial image from expression neutrality, mapped to a quality measure fully compliant with ISO/IEC 29794-5.
    \item We evaluate the performance of our NeutrEx measure and benchmark it against baseline candidate algorithms of the \textit{Open Source Face Image Quality} (OFIQ) framework\footnote{\url{https://github.com/BSI-OFIQ/OFIQ-Project}} measures for expression neutrality. We assess the performance on two tasks: \textit{expression neutrality classification} and \textit{FR Utility Prediction}. We compare the results using two well-known datasets \cite{Gross-MultiPIE-IVC-2010, GAN-FEAFA+-ICDIP-2022}, as well as a commercial off-the-shelf (COTS) system\footnote{\url{https://www.cognitec.com/} (FaceVACS, Version 9.6.0)} and an open-source FR system \cite{Meng-MagFace-CVPR-2021}.
\end{itemize}

This paper is structured as follows: First, Section~\ref{sec:monocular-face-reconstruction} explains the foundational concepts behind \textit{Monocular 3D Face Reconstruction}. Next, Section~\ref{sec:neutrex} gives a more detailed overview of our proposed NeutrEx measures along with the baseline approaches in Section~\ref{sec:baselines}. The evaluation and discussion of limitations are presented in Section~\ref{sec:evaluation} and Section~\ref{sec:discussion}, followed by a final conclusion and future outline in Section~\ref{sec:conclusion}.

\section{Monocular Face Reconstruction}
\label{sec:monocular-face-reconstruction}

A 3D Morphable Face Model is defined as a generative model based on two essential concepts~\cite{Egger-3DMM-Survey-TOG-2020}: On the one hand, all face images projected into model space adhere to a dense vertex-to-vertex correspondence. In this work, we exploit this property by computing \textit{per-vertex} Euclidean Distances between two 3D face meshes to quantify their dissimilarity. On the other hand, 3DMMs divide subject-related attributes, such as facial shape, expression, or skin colour and disentangle them from external attributes like illumination or camera settings.    

The current state-of-the-art 3DMM proposed by Li et al.~\cite{Li-FLAME-ToG-2017}, namely \textit{FLAME}, consists of 5,023 vertices, including four joints positioned at jaw, neck, and eyes. The FLAME decoder $F$ maps a set of low-dimensional parameters to a set of vertices V: $F(\beta, \theta, \psi) \mapsto V\in\mathbb{R}^{5023\times3}$ to control head shape $\beta$, head pose $\theta$, and facial expression $\psi$, respectively. 

In this context, Monocular 3D Face Reconstruction aims to project a single 2D face image into the 3DMM parameter space. Translated to FLAME, this means finding the respective parameters $\beta$, $\theta$, and $\psi$ that correspond to the 3D face structure underlying the 2D face image. Addressing this issue, DECA~\cite{Feng-DECA-ToG-2021} constitutes a milestone as it trains a \textit{coarse} and \textit{detail} encoder~\cite{He-Resnet-CVPR-2016} to encode face images into the FLAME parameter space. The authors introduce a \textit{consistency} loss term employed during training to disentangle identity-specific details from expression-related wrinkles. Recently, EMOCA\cite{Danvevcek-EMOCA-CVPR-2022} was presented as a follow-up work that incorporates an emotion consistency loss to reconstruct facial images covering a wider range of emotions, including more subtle facial expressions. 

In this work, we use the DECA coarse shape encoder in combination with the emotion-sensitive expression encoder from EMOCA to predict $\beta_i$, $\theta_i$, and $\psi_i$ for each face image $I_i$ in the evaluation dataset. A more detailed description of our methodology follows in Section~\ref{sec:methodology}.

\begin{figure*}
\centering
\includegraphics[width=0.8\linewidth]{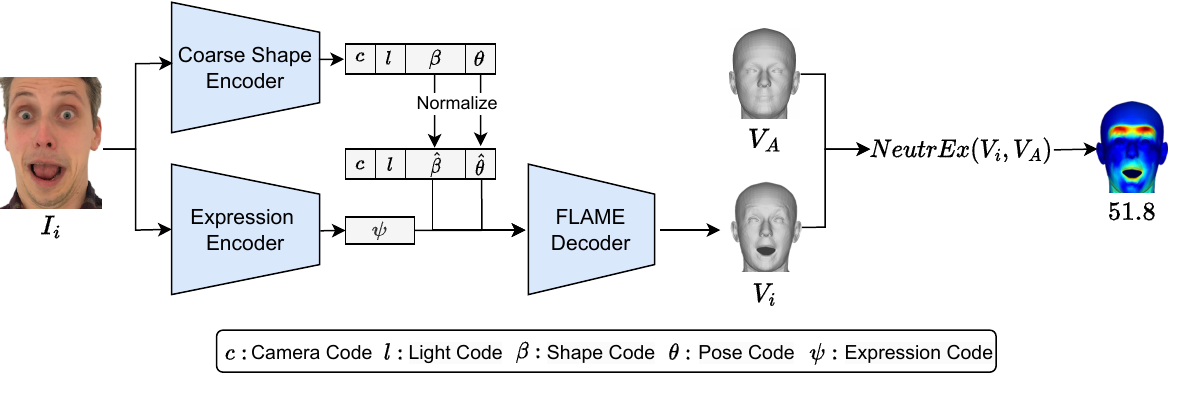}
\caption{Underlying architecture for computing the NeutrEx measure based on FLAME~\cite{Li-FLAME-ToG-2017} 3D face reconstructions. Facial images are encoded into the FLAME parameter space using the coarse shape encoder from DECA~\cite{Feng-DECA-ToG-2021} and the expression encoder of EMOCA~\cite{Danvevcek-EMOCA-CVPR-2022}.}
\label{fig:emoca-overview}
\end{figure*}

\section{Methodology}
\label{sec:methodology}

In this section, we present our NeutrEx measure and the baseline approaches established to validate the effectiveness of our results. 

\subsection{NeutrEx Measure}
\label{sec:neutrex}

Figure~\ref{fig:emoca-overview} provides an overview of the 3D face reconstruction process underlying the computation of our NeutrEx measures. We denote $V_i\in\mathbb{R}^{5,023\times3}$ as the reconstruction of face image $I_i$ given shape code $\beta_i$, pose code $\theta_i$, and expression code $\psi_i$. As the light condition $l$ and the distance to the camera $c$ have no influence on the vertex positions, we ignore both parameters in the calculation of our NeutrEx measures. Prior to the further processing of $V_i$, we emphasize the importance of transforming each 3D face model into a generic head shape ($\hat{\beta}=0$) that faces the camera with a frontal view ($\hat{\theta}=0$). This ensures that any Euclidean distances measured between the vertices of two 3D face models solely focus on the \textit{impact of facial expressions}, separating it from any \textit{identity-specific} or \textit{pose-specific} variation.

To calculate the distance between individual facial expressions to expression neutrality, it is essential to establish a neutral anchor that represents the average neutral expression (see Figure~\ref{fig:class-avg-residuals}). Here, we obtain the neutral anchor $V_A$ by computing the average expression code $\psi_{A}$ upon all neutral expressions in the training datasets~\cite{Gross-MultiPIE-IVC-2010}\cite{GAN-FEAFA+-ICDIP-2022}. We define the per-vertex L2 distance between $V_i$ and $V_A$ as follows:

\begin{equation}
    D(V_i) = ||V_i - V_A||_2
\label{eq:neutrex-diffs}
\end{equation}

To convert the L2 distances into quality component values within the range of $[0, 100]$ according to ISO/IEC 29794-5, we apply \textit{min-max scaling} with $D_{min}$ and $D_{max}$ empirically determined from the training datasets:

\begin{equation}
    \mathit{NeutrEx}(V_i) = 100 \cdot\left(1 - \frac{D(V_i) - D_{\min}}{D_{\max} - D_{\min}}\right)
    \label{eq:neutrex-quality}
\end{equation}

During the evaluation, we handle accumulated distances falling below $D_{min}$ or exceeding $D_{max}$ by clipping $\mathit{NeutrEx}(V_i)$ within the range of 0 to 100.

\subsection{Baseline Measures}
\label{sec:baselines}

\begin{figure}
\centering
\includegraphics[width=\linewidth]{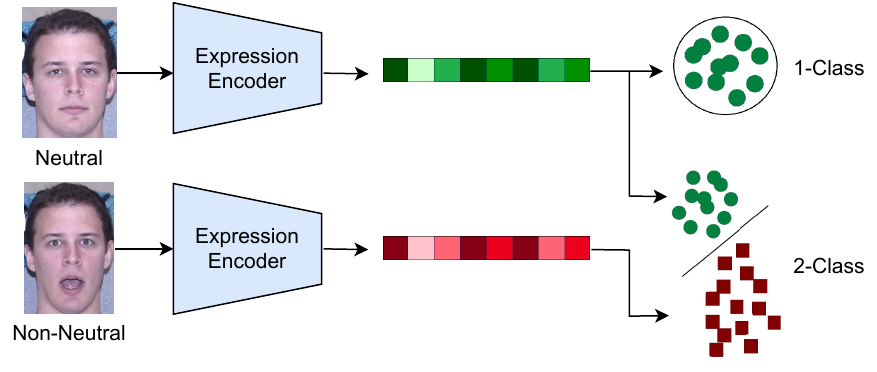}
\caption{Overview of the extraction of neutral and non-neutral face embeddings based on the baseline DMUE~\cite{She-DMUE-CVPR-2021} facial expression classifier. Our Baseline measures are established by training one-class and two-class SVMs.}
\label{fig:DMUE-overview}
\end{figure}

To demonstrate the effectiveness of our NeutrEx measure, we benchmark with two baseline candidate measures of the \textit{OFIQ framework}\footnote{\url{https://github.com/BSI-OFIQ/OFIQ-Project}} for facial expression neutrality, the general approaches of which are shown in Figure \ref{fig:DMUE-overview}. Similar to the NIST Fingerprint Image Quality 2 (NFIQ2) framework~\cite{NIST-NFIQ2-FingerprintImageQuality-2021}, OFIQ is planned to be an open source framework that links face image (component) quality to operational recognition performance. 

\begin{figure*}
\centering
  \begin{subfigure}[b]{0.13\linewidth}
    \includegraphics[width=\linewidth]{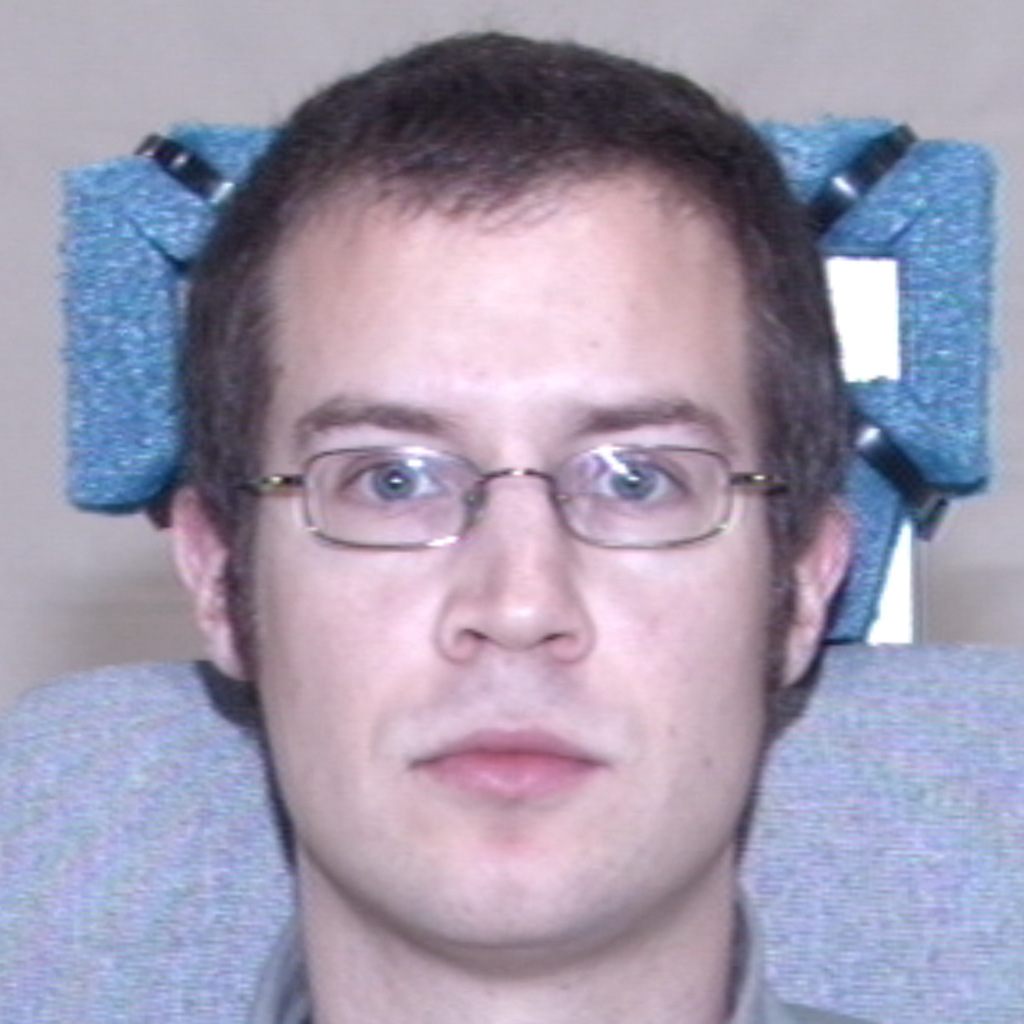}
    \label{fig:synth-face-example}
  \end{subfigure} \hspace{15pt}
    \begin{subfigure}[b]{0.13\linewidth}
    \includegraphics[width=\linewidth]{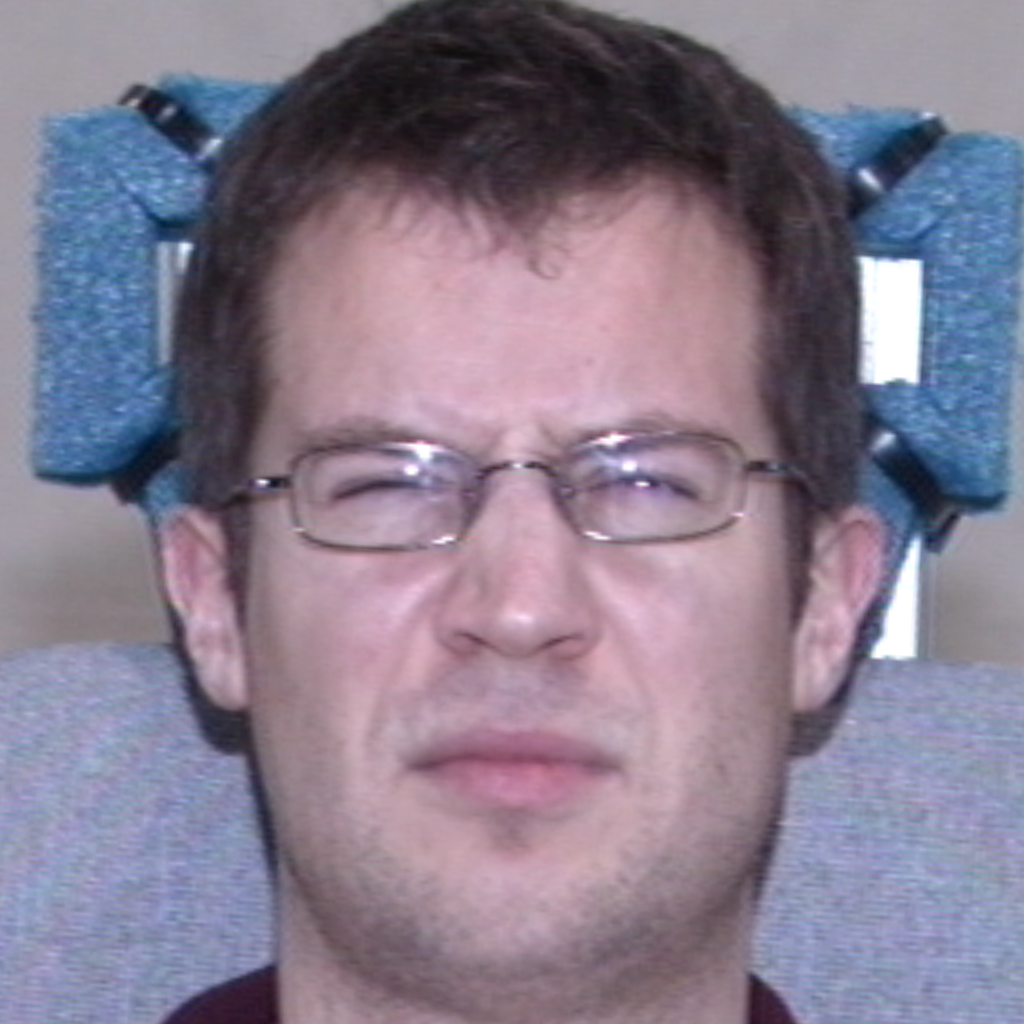}
    \label{fig:synth-face-example}
  \end{subfigure}
    \begin{subfigure}[b]{0.13\linewidth}
    \includegraphics[width=\linewidth]{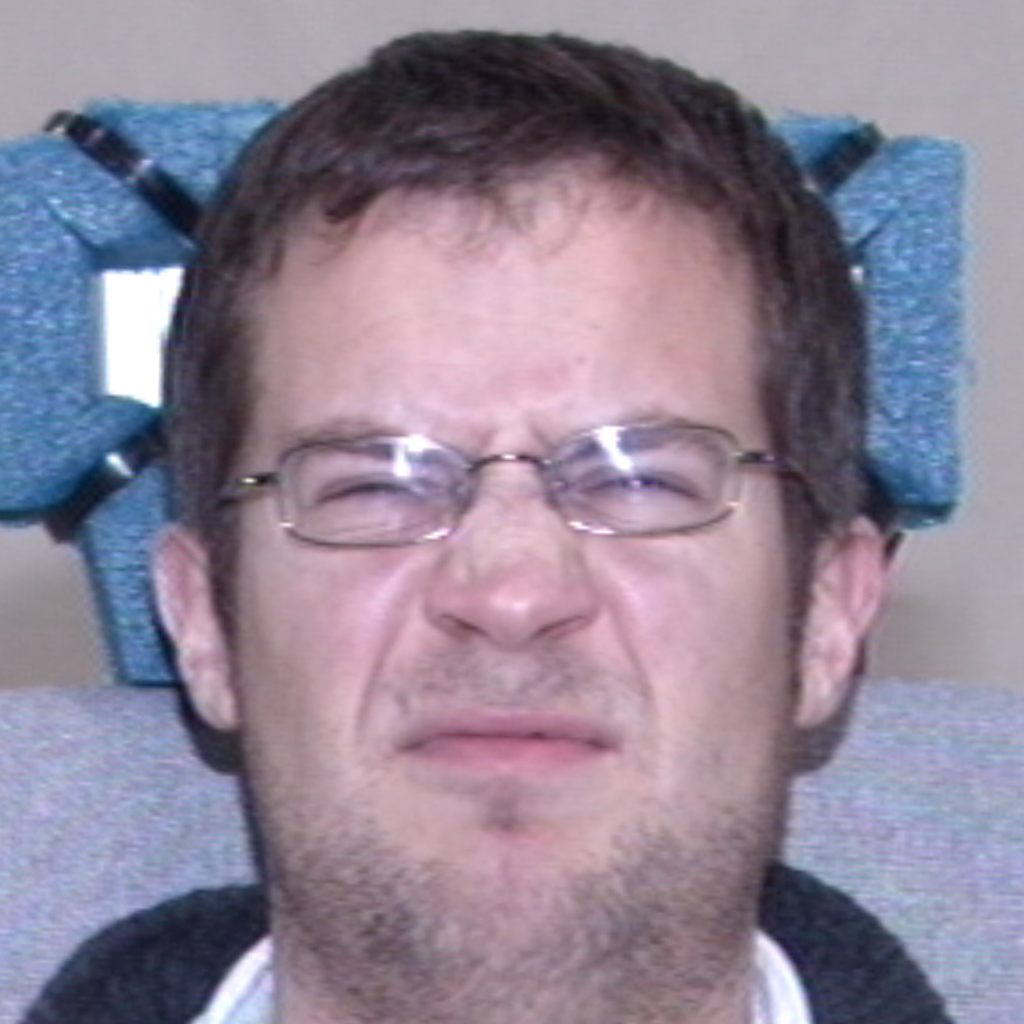}
    \label{fig:real-face-example}
  \end{subfigure}
  \begin{subfigure}[b]{0.13\linewidth}
    \includegraphics[width=\linewidth]{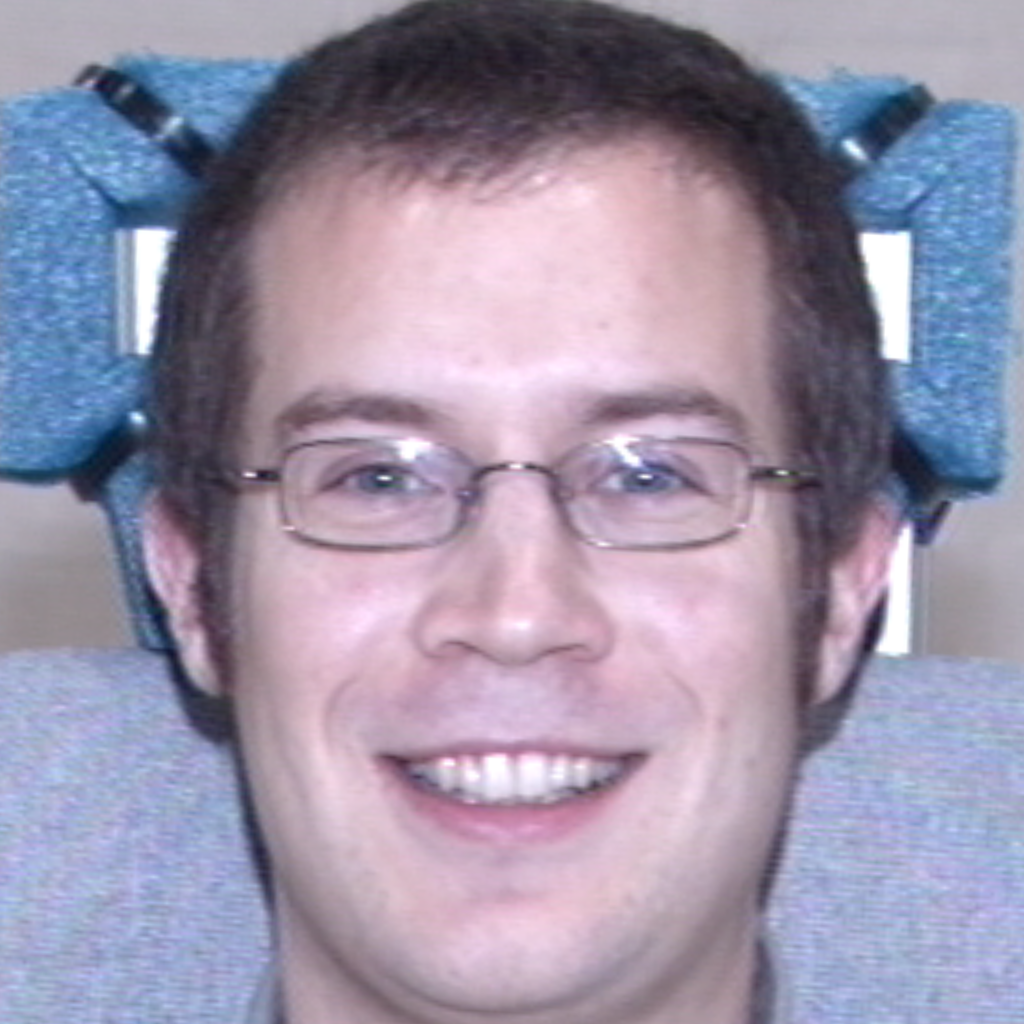}
    \label{fig:real-face-example}
  \end{subfigure}
  \begin{subfigure}[b]{0.13\linewidth}
    \includegraphics[width=\linewidth]{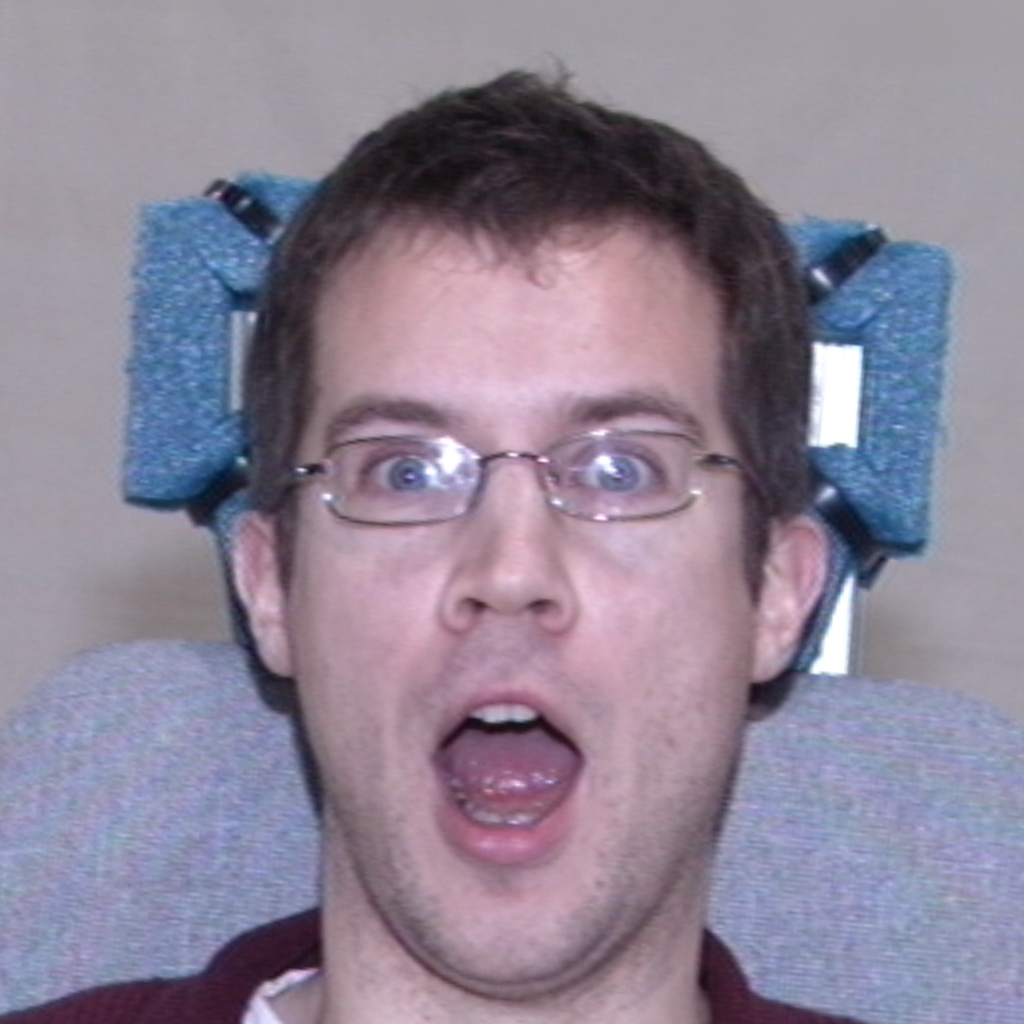}
    \label{fig:real-face-example}
  \end{subfigure}
  \begin{subfigure}[b]{0.13\linewidth}
    \includegraphics[width=\linewidth]{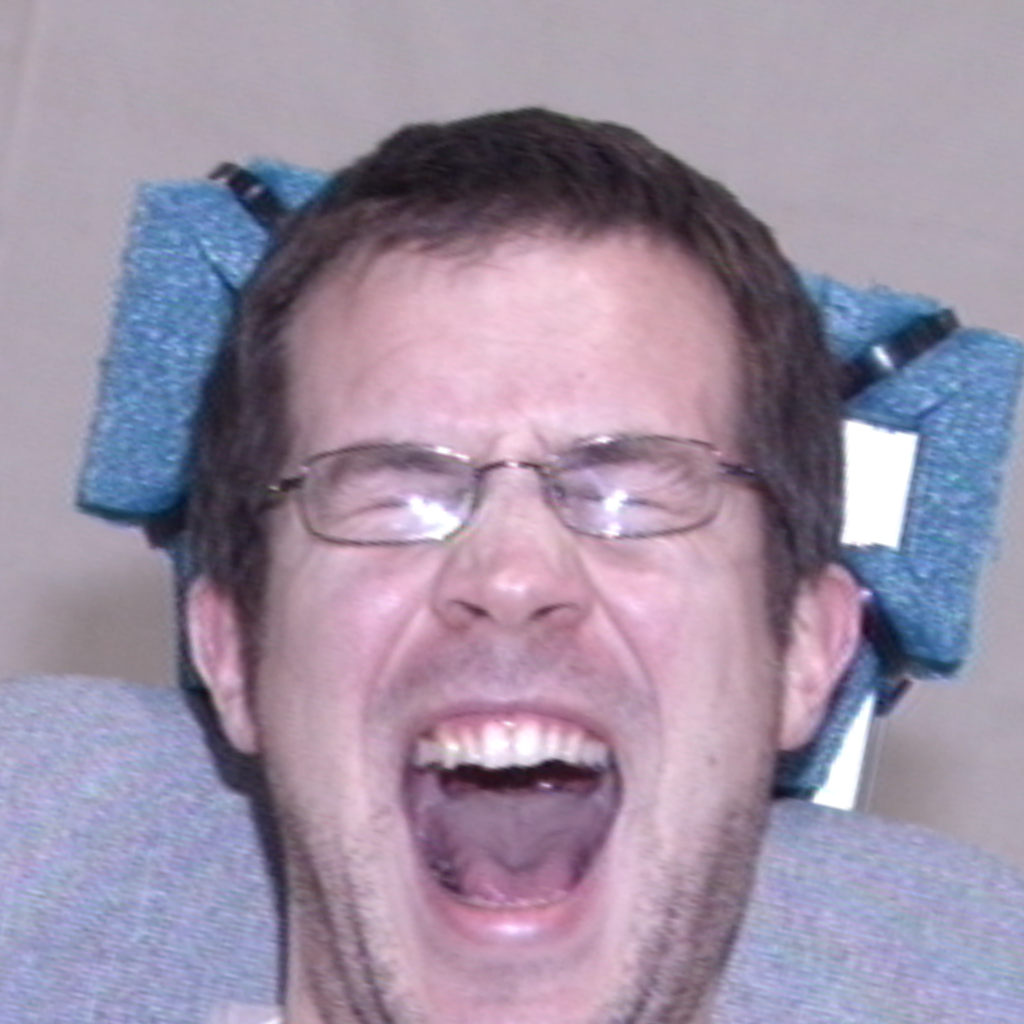}
    \label{fig:real-face-example}
  \end{subfigure}
  
  \begin{subfigure}[b]{0.13\linewidth}
    \includegraphics[width=\linewidth]{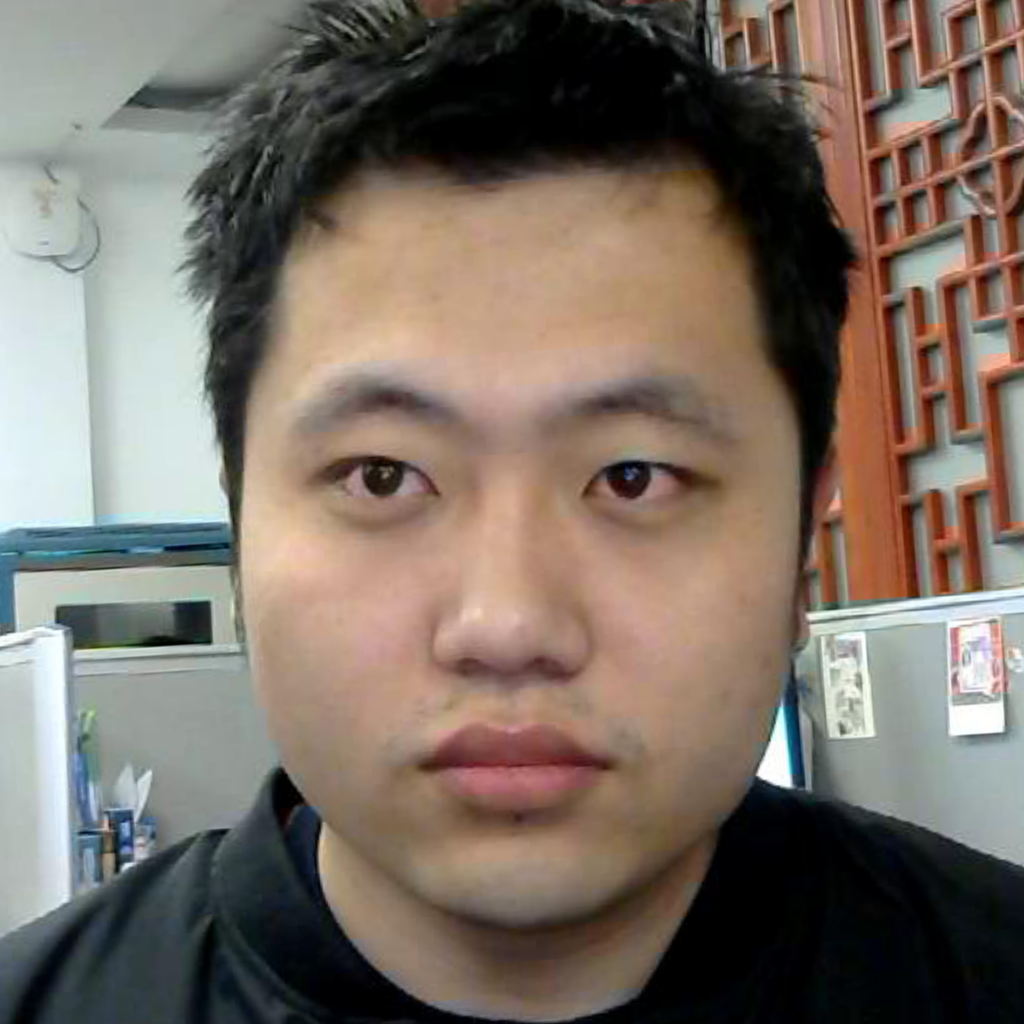}
    \caption{Neutral}
    \label{fig:synth-face-example}
  \end{subfigure}  \hspace{15pt}
    \begin{subfigure}[b]{0.13\linewidth}
    \includegraphics[width=\linewidth]{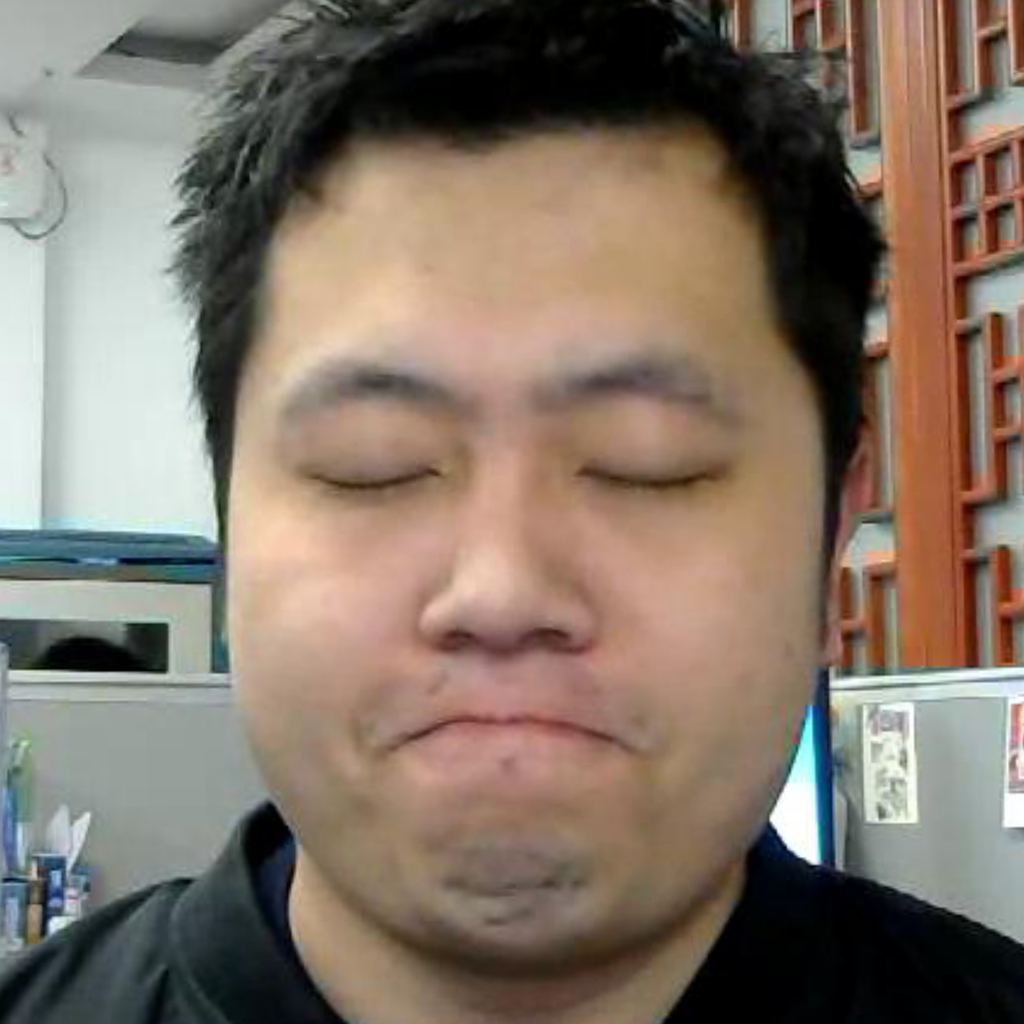}
    \caption*{ }
    \label{fig:synth-face-example}
  \end{subfigure}
    \begin{subfigure}[b]{0.13\linewidth}
    \includegraphics[width=\linewidth]{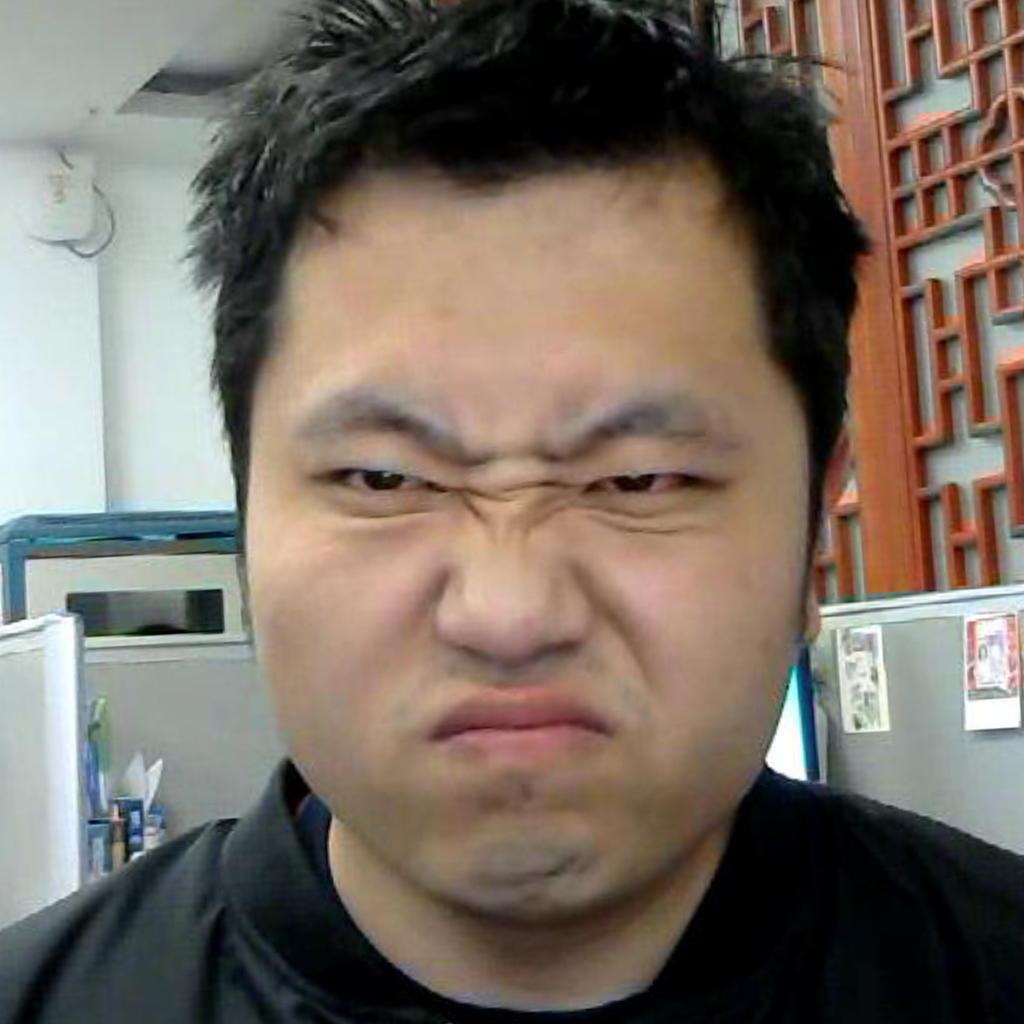}
    \caption*{ }
    \label{fig:real-face-example}
  \end{subfigure}
  \begin{subfigure}[b]{0.13\linewidth}
    \includegraphics[width=\linewidth]{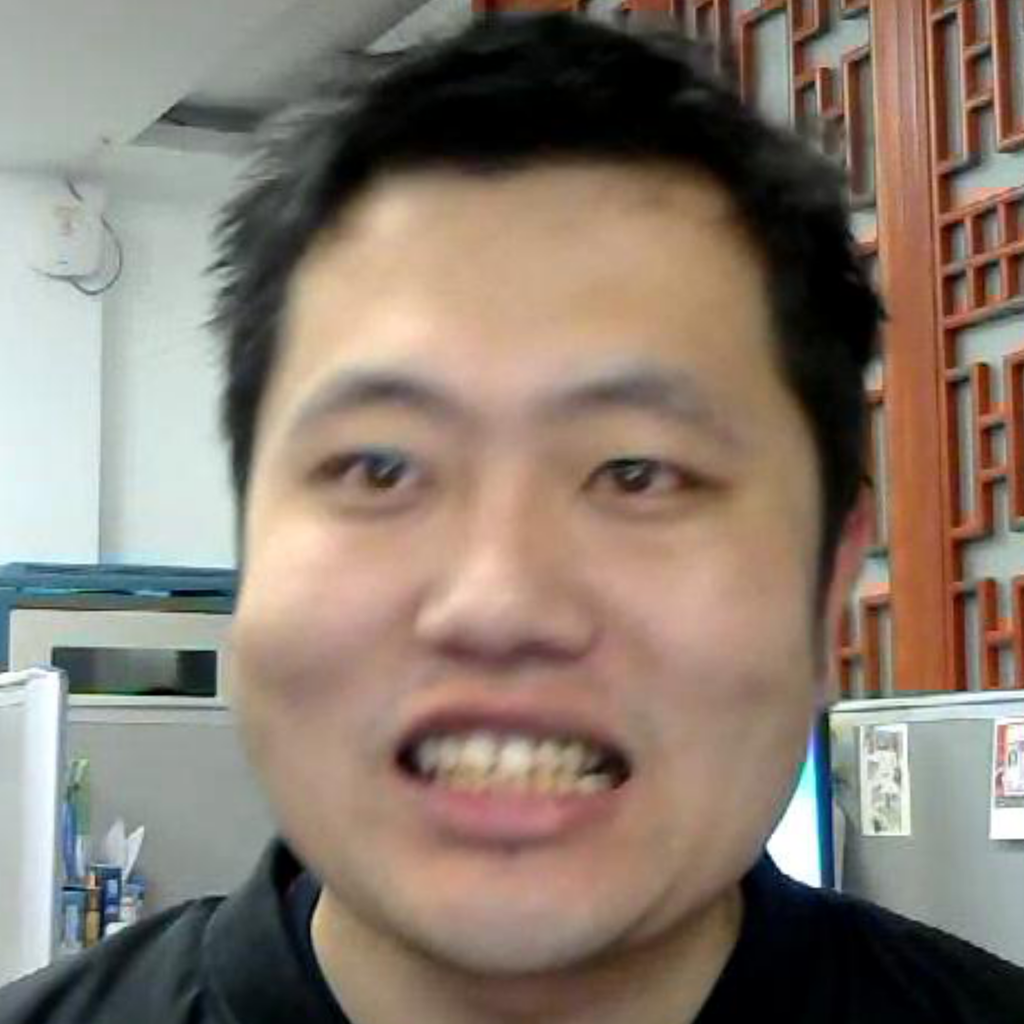}
    \caption{Non-Neutral}
    \label{fig:real-face-example}
  \end{subfigure}
  \begin{subfigure}[b]{0.13\linewidth}
    \includegraphics[width=\linewidth]{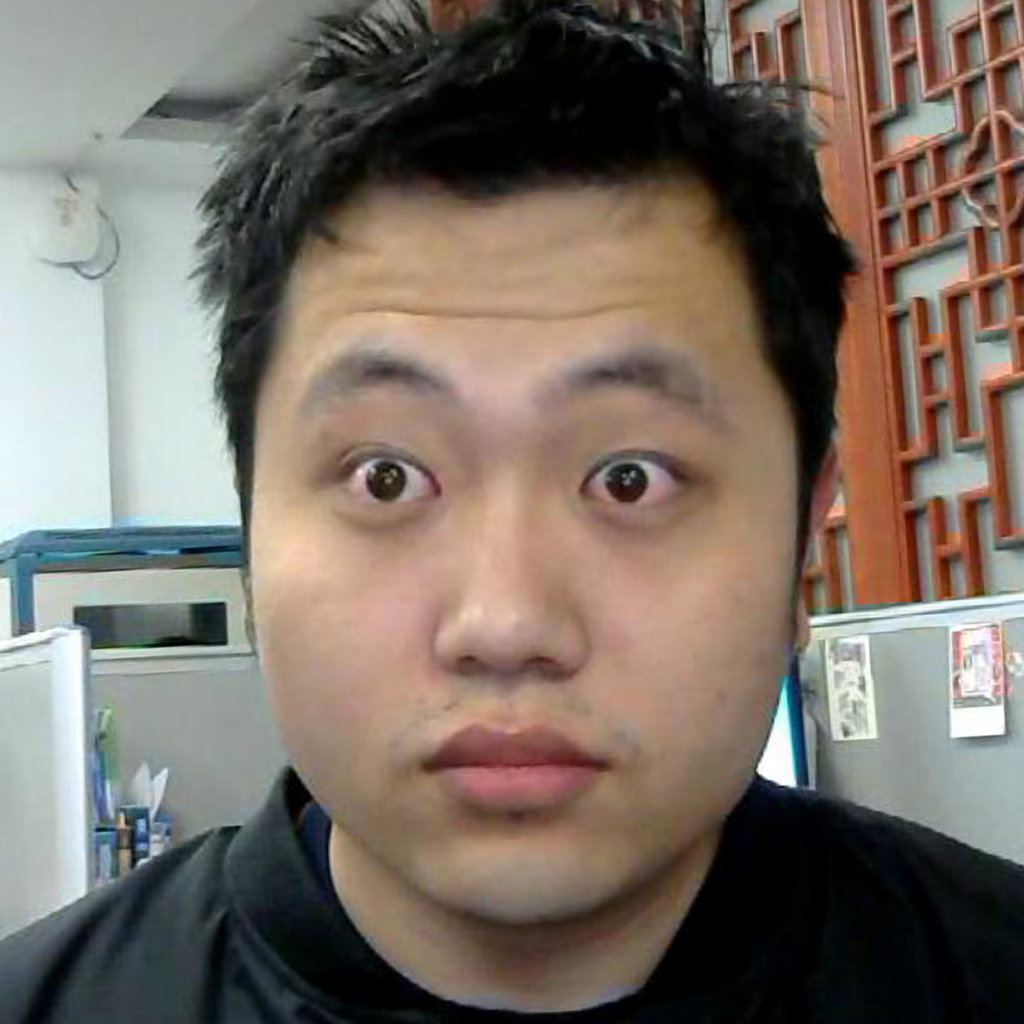}
    \caption*{ }
    \label{fig:real-face-example}
  \end{subfigure}
  \begin{subfigure}[b]{0.13\linewidth}
    \includegraphics[width=\linewidth]{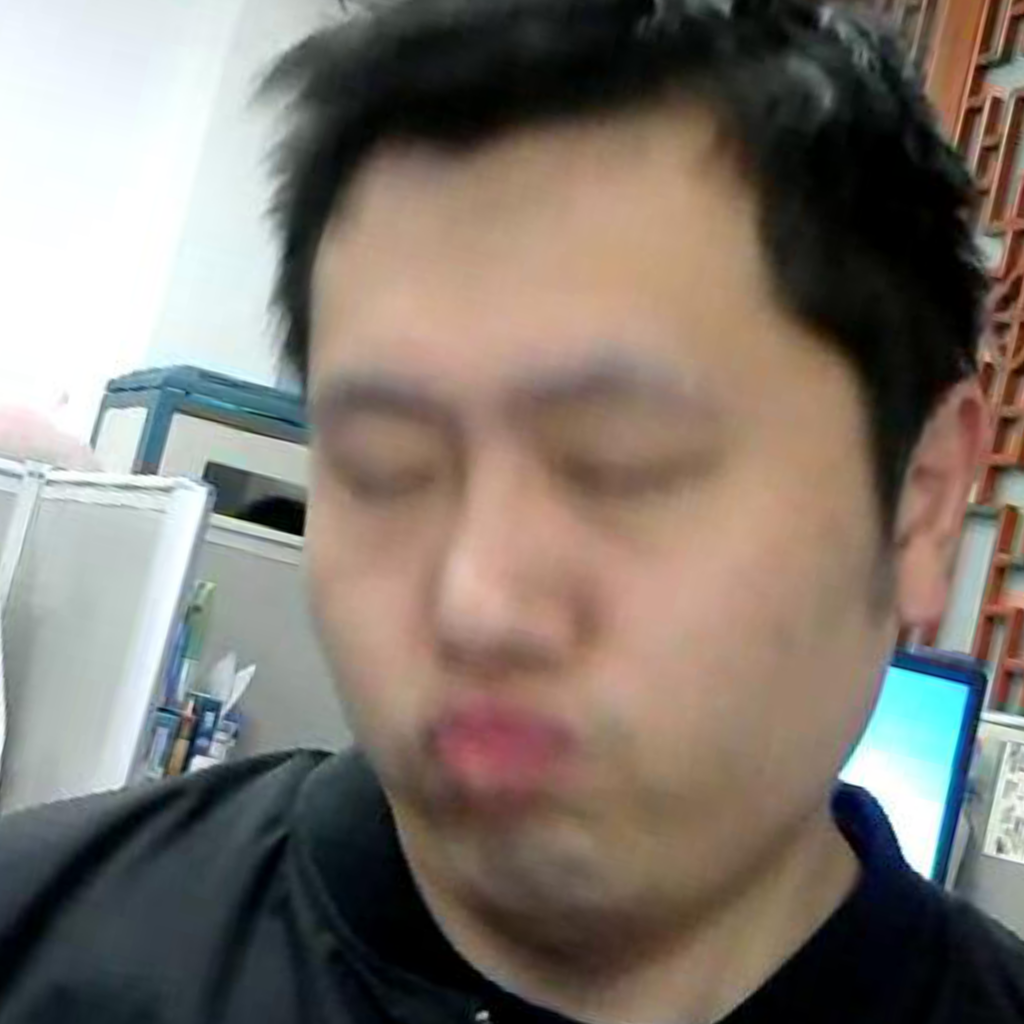}
    \caption*{ }
    \label{fig:real-face-example}
  \end{subfigure}
  \caption{Example face images representing each expression class in the Multi-PIE dataset~\cite{Gross-MultiPIE-IVC-2010} (top). Randomly selected neutral and non-neutral samples from the FEAFA+ dataset~\cite{GAN-FEAFA+-ICDIP-2022} (bottom). \label{fig:multi-pie-examples}}
\end{figure*}

\begin{table*}
\caption{Datasets used to evaluate our NeutrEx measures include a constraint dataset (Multi-PIE~\cite{Gross-MultiPIE-IVC-2010}) with variation in facial expression only and a semi-constraint dataset (FEAFA+~\cite{GAN-FEAFA+-ICDIP-2022}) that contains additional variations in motion blur and pose angles. 
\label{tab:multi-pie-sample-statistics}}
\centering
\begin{tabular}{|l|c|c|c|cllll|c|}
\hline
\textbf{Dataset} & \textbf{IDs} & \textbf{Images} & \textbf{Neutral Samples} & \multicolumn{5}{c|}{\textbf{Non-Neutral Samples}} & \textbf{Mated Comparisons}  \\ \hline
Multi-PIE~\cite{Gross-MultiPIE-IVC-2010}        & 337          & 2,512           & 1,158                    & \multicolumn{5}{c|}{1,354}                        & 2,175                                      \\ \hline
FEAFA+~\cite{GAN-FEAFA+-ICDIP-2022}           & 117          & 41,339          & 15,462                   & \multicolumn{5}{c|}{25,877}                       & 41,222                                  \\ \hline
\end{tabular}
\end{table*}

Technically, we extract the face embeddings of a Convolutional Neural Network that has been pre-trained on facial expression classification. Given these face embeddings, we train \textit{one-class} and \textit{two-class Support Vector Machines} (SVMs) to classify between \textit{neutral} vs. \textit{non-neutral} facial expressions. During inference, we use the distance of each face embedding to the SVM decision boundary of the neutral cluster to obtain the final baseline measures. 

Specifically, we utilize the pre-trained DMUE classifier of She et al.~\cite{She-DMUE-CVPR-2021} based on a Resnet-18~\cite{He-Resnet-CVPR-2016} backbone to predict facial expressions among the following classes: Neutral, Happy, Disgust, Surprised, Contempt, Angry, and Sad.

First, we discard the 7-dimensional output layer of the final DMUE fully connected neural network, arguing that the preceding 512-dimensional hidden layer (\textit{face embedding)} includes more distinctive features relevant to the classification of facial expressions. After extracting all face embeddings from the training datasets~\cite{Gross-MultiPIE-IVC-2010}\cite{GAN-FEAFA+-ICDIP-2022}, we train a \textit{one-class} SVM\footnote{SVMs trained with Scikit-Learn Version 1.2.1 using an \textit{rbf-kernel} with $\nu = 0.05$} only on the neutral face embeddings. Consequently, non-neutral facial expressions during inference are treated as \textit{anomaly} and assigned an expression neutrality score reflecting the side and distance to the learned hyperplane of the neutral embedding class. Following Equation~\ref{eq:neutrex-quality}, each distance is converted into an ISO/IEC-compliant quality component value.

Additionally, we use a \textit{two-class} SVM to classify the face embeddings into \textit{neutral} and \textit{non-neutral} expressions. Unlike the one-class approach, this strategy learns from all expression classes in the training dataset.

\section{Evaluation}
\label{sec:evaluation}

In this section, we cover the dataset selection and present the performance comparison between our NeutrEx and baseline measures and their ability to provide explainable results.

\subsection{Datasets}
\label{subsec:datasets}

We base our experiments on the Multi-PIE~\cite{Gross-MultiPIE-IVC-2010} and FEAFA+~\cite{GAN-FEAFA+-ICDIP-2022} datasets with key details summarised in Table~\ref{tab:multi-pie-sample-statistics} and examples of neutral and non-neutral images shown in Figure~\ref{fig:multi-pie-examples}. Multi-PIE comprises facial images captured under controlled conditions, including uniform illumination and frontal views only. In contrast, FEAFA+ encompasses a wide range of facial expressions but also contains other types of variation, such as motion blur and diverse pose angles. To assess the performance independent of training data, we compute the NeutrEx neutral anchor and baseline SVMs separately using Multi-PIE and FEAFA+. This \textit{leave-one-out} training-evaluation scheme allows evaluating the cross-database performance and the measures' ability to generalize and handle facial expressions not seen during training.  

\begin{figure}[h]
\centering
\begin{subfigure}{0.49\linewidth}
    \centering
    \includegraphics[width=\linewidth]{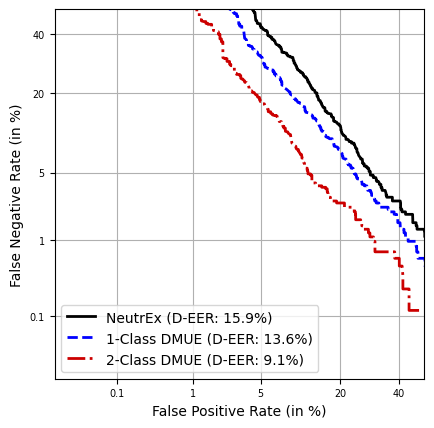}
    \caption{Multi-PIE}
\end{subfigure}
\begin{subfigure}{0.49\linewidth}
    \centering
    \includegraphics[width=\linewidth]{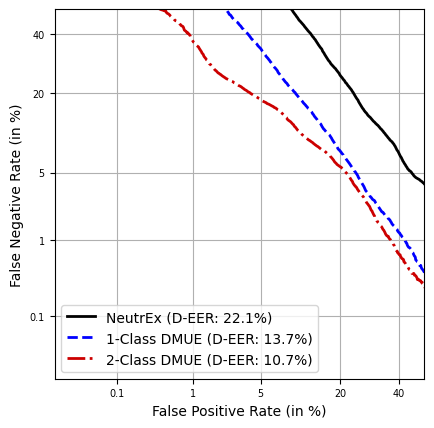}
    \caption{FEAFA+}
\end{subfigure}
\caption{DET curves showing the performance evaluation of our NeutrEx measure and baseline measures for classifying neutral vs. non-neutral expressions, along with their corresponding D-EERs.}
\label{fig:stacked-area-plots}
\end{figure}

\subsection{Expression Neutrality Classification}
\label{sec:expression-neutrality-estimation}

\begin{figure*}
\centering
\begin{subfigure}{0.32\linewidth}
    \centering
    \includegraphics[width=\linewidth]{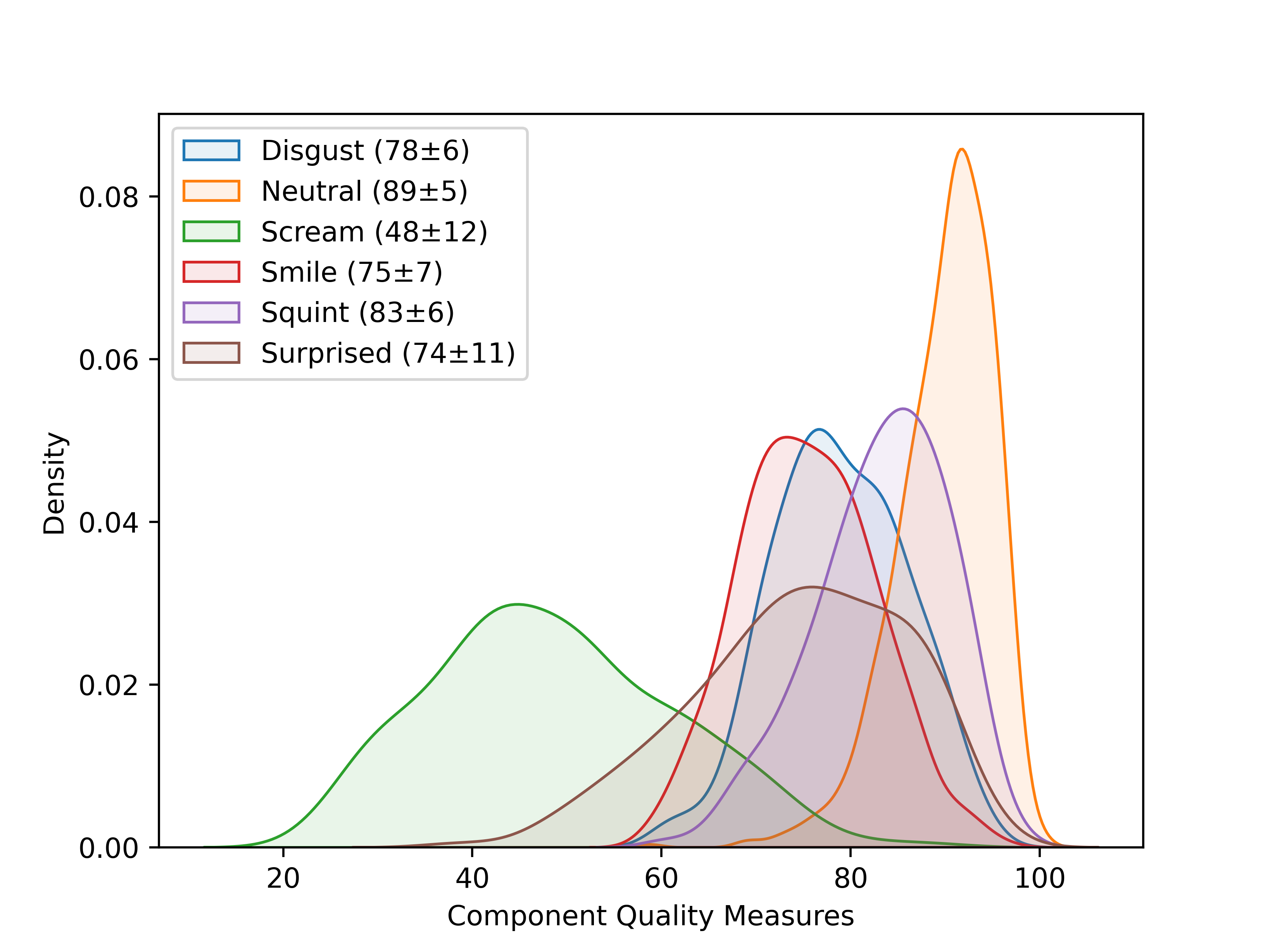}
    \caption{NeutrEx}
\end{subfigure}
\begin{subfigure}{0.32\linewidth}
    \centering
    \includegraphics[width=\linewidth]{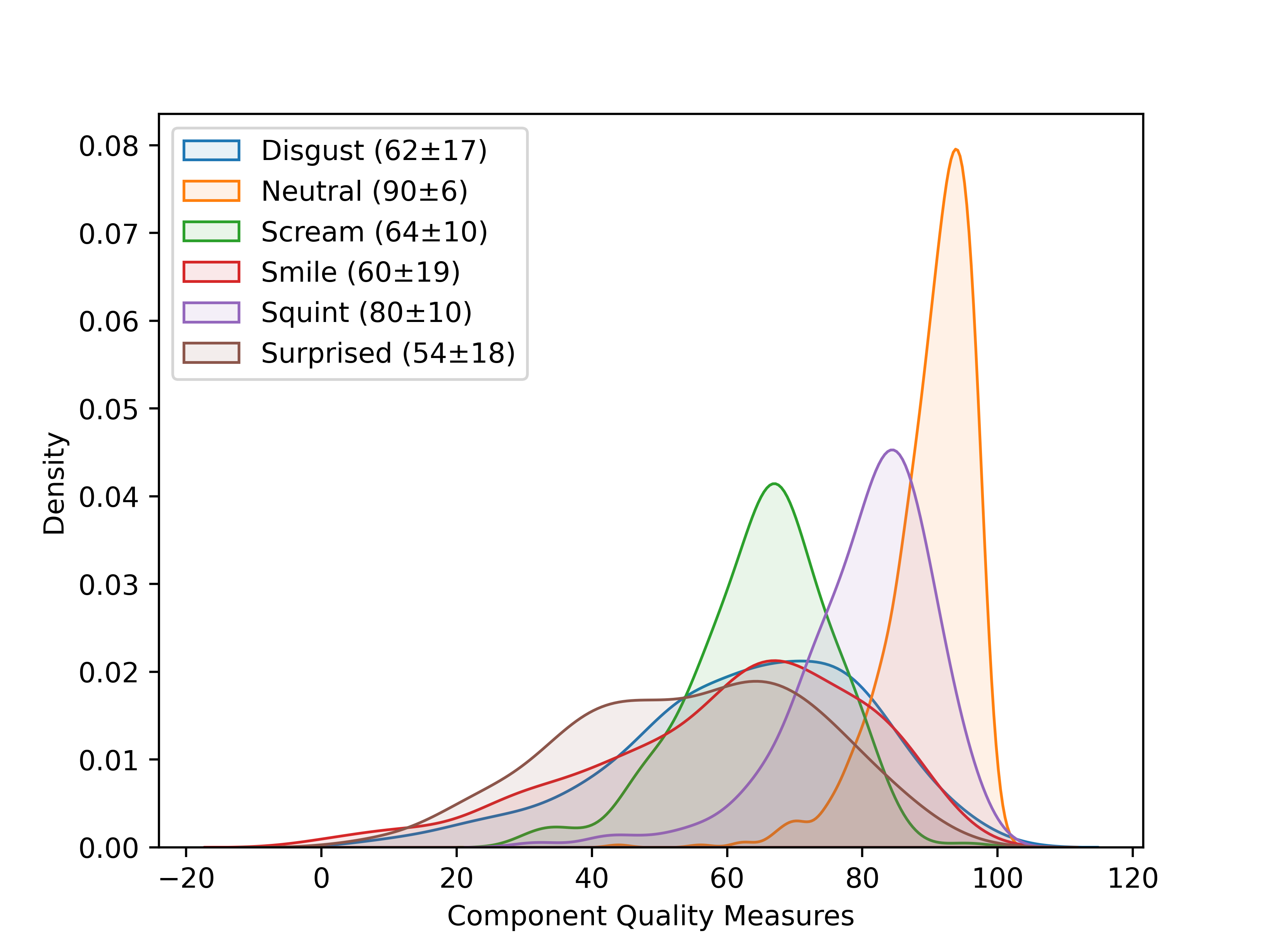}
    \caption{DMUE (1-Class)}
\end{subfigure}
\begin{subfigure}{0.32\linewidth}
    \centering
    \includegraphics[width=\linewidth]{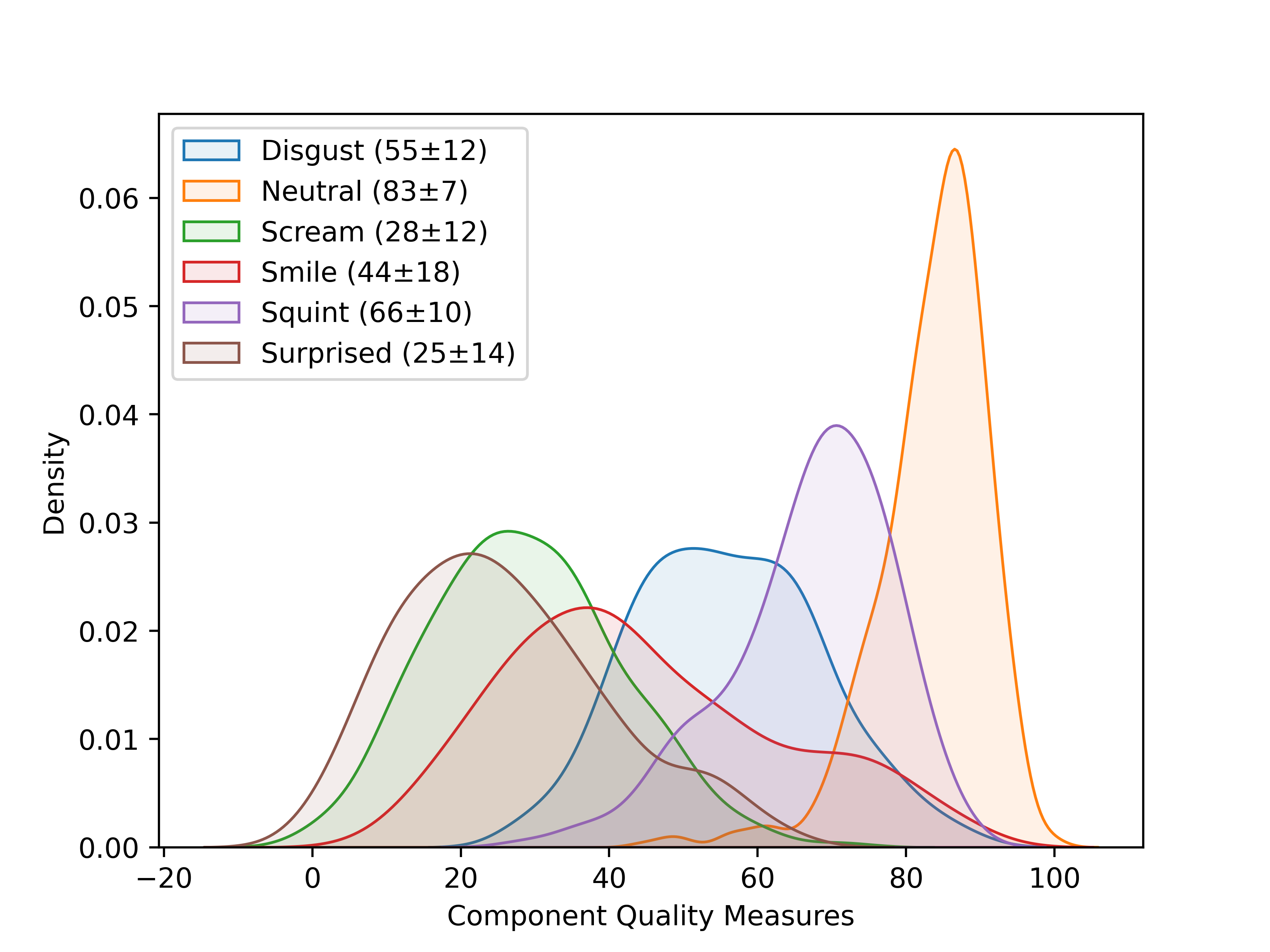}
    \caption{DMUE (2-Class)}
\end{subfigure}
\caption{Class-wise distributions of the NeutrEx and baselines measures with their corresponding means and standard deviations based on the Multi-PIE dataset~\cite{Gross-MultiPIE-IVC-2010}.}
\label{fig:stacked-area-plots}
\end{figure*}

\begin{figure*}
  \centering
  \begin{tabular}{ccccccc}
    \textbf{Reference} & \textbf{Neutral} & \textbf{Squint} & \textbf{Disgust} & \textbf{Smile} & \textbf{Surprised} & \textbf{Scream} \\[-1pt]
     & ($90.0$) & ($83.3$) & ($78.8$) & ($75.4$) & ($74.5$) & ($48.5$) \\[10pt]
    \adjustbox{valign=c}{\includegraphics[width=0.12\textwidth]{images/neutral-reference.png}} &
    \includegraphics[width=0.12\textwidth]{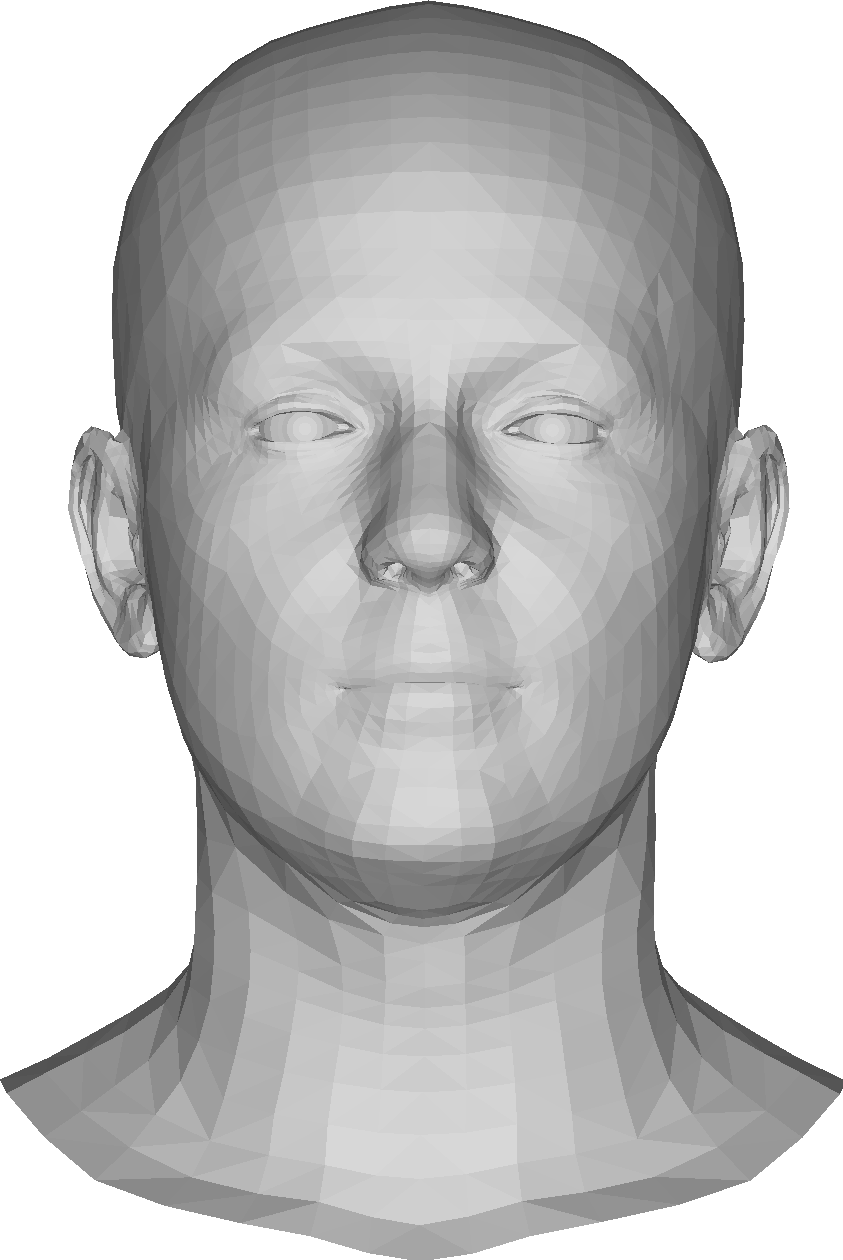} &
    \includegraphics[width=0.12\textwidth]{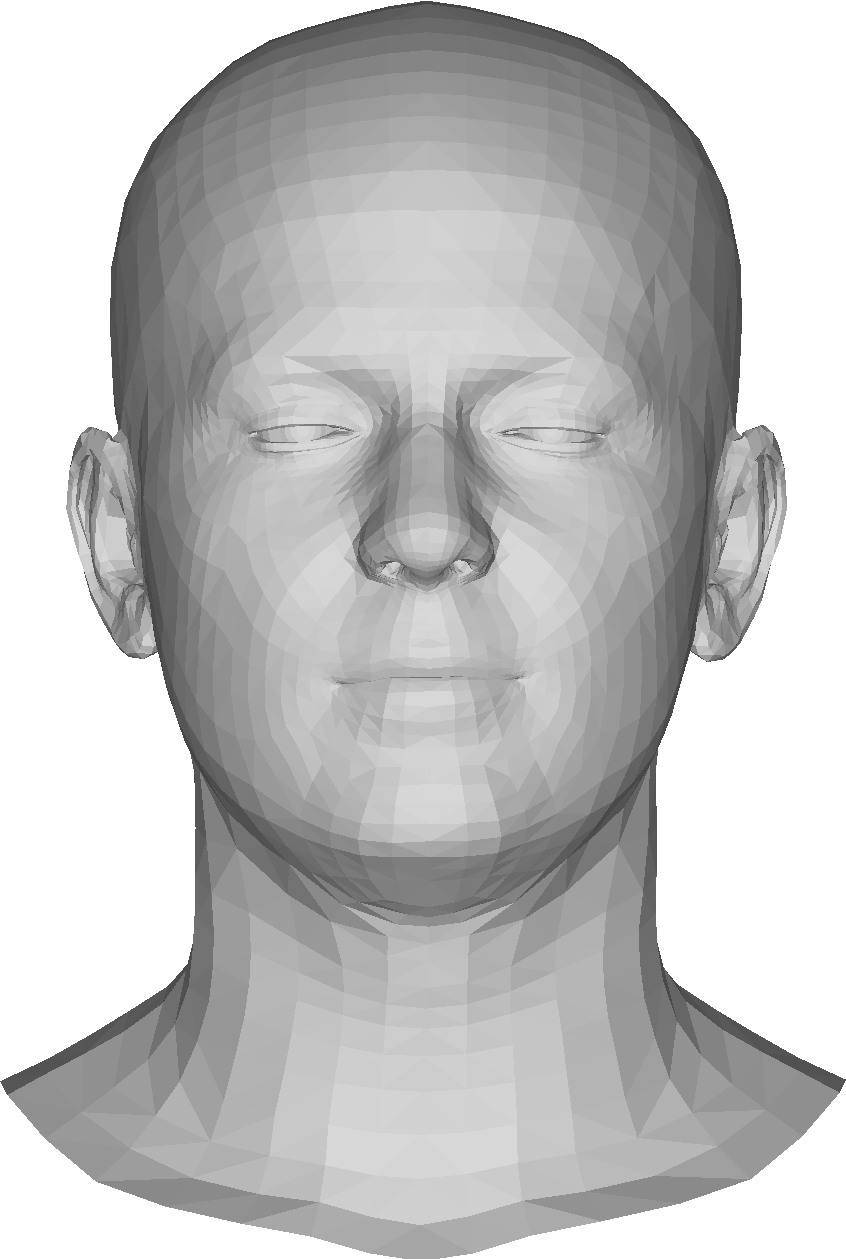} &
    \includegraphics[width=0.12\textwidth]{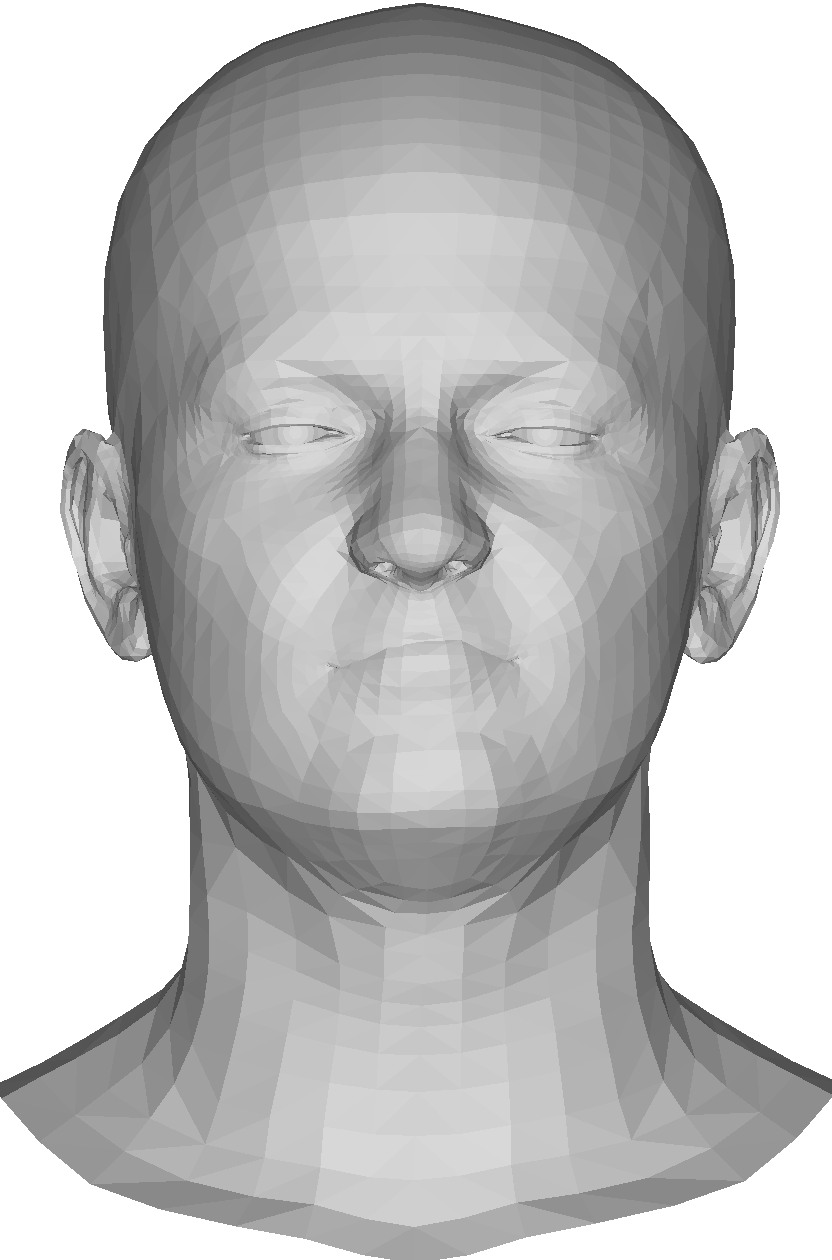} &
    \includegraphics[width=0.12\textwidth]{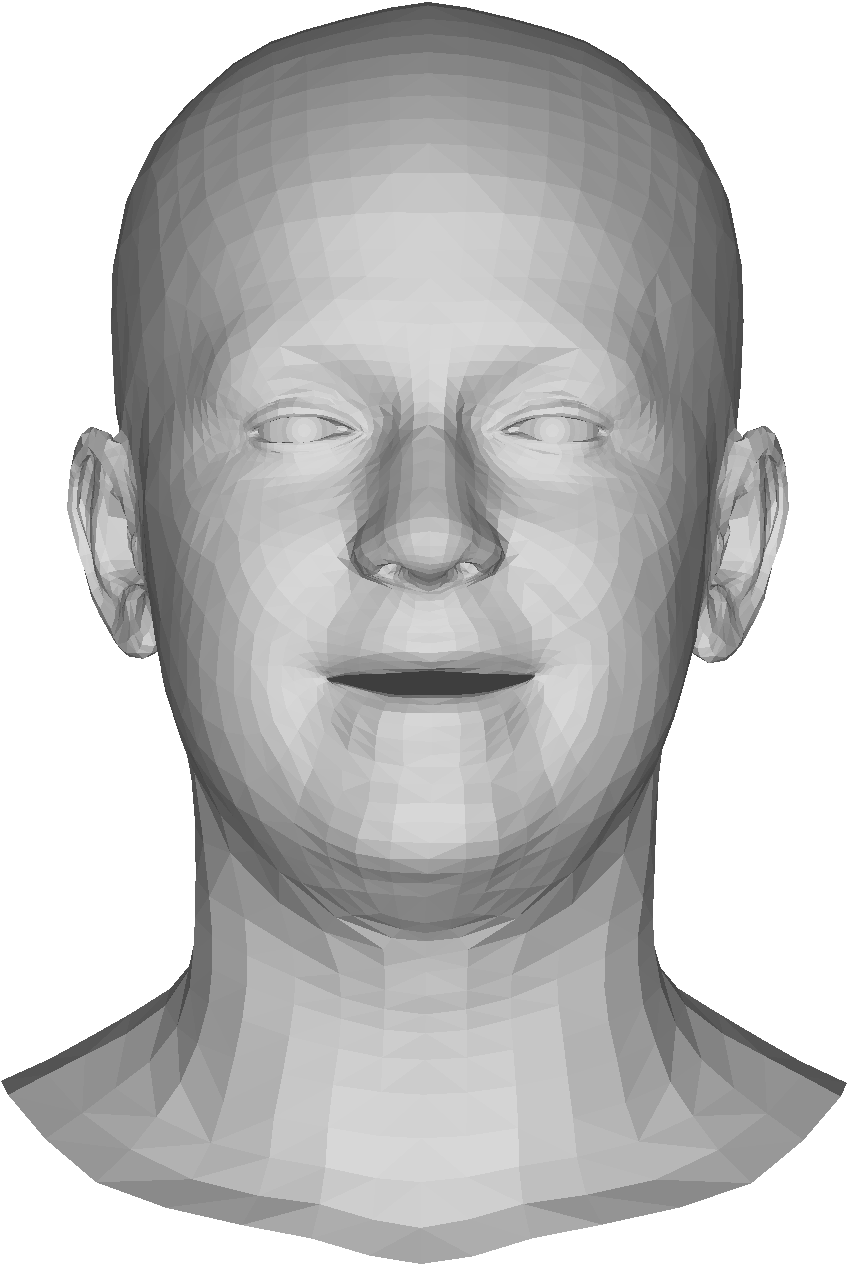} &
    \includegraphics[width=0.12\textwidth]{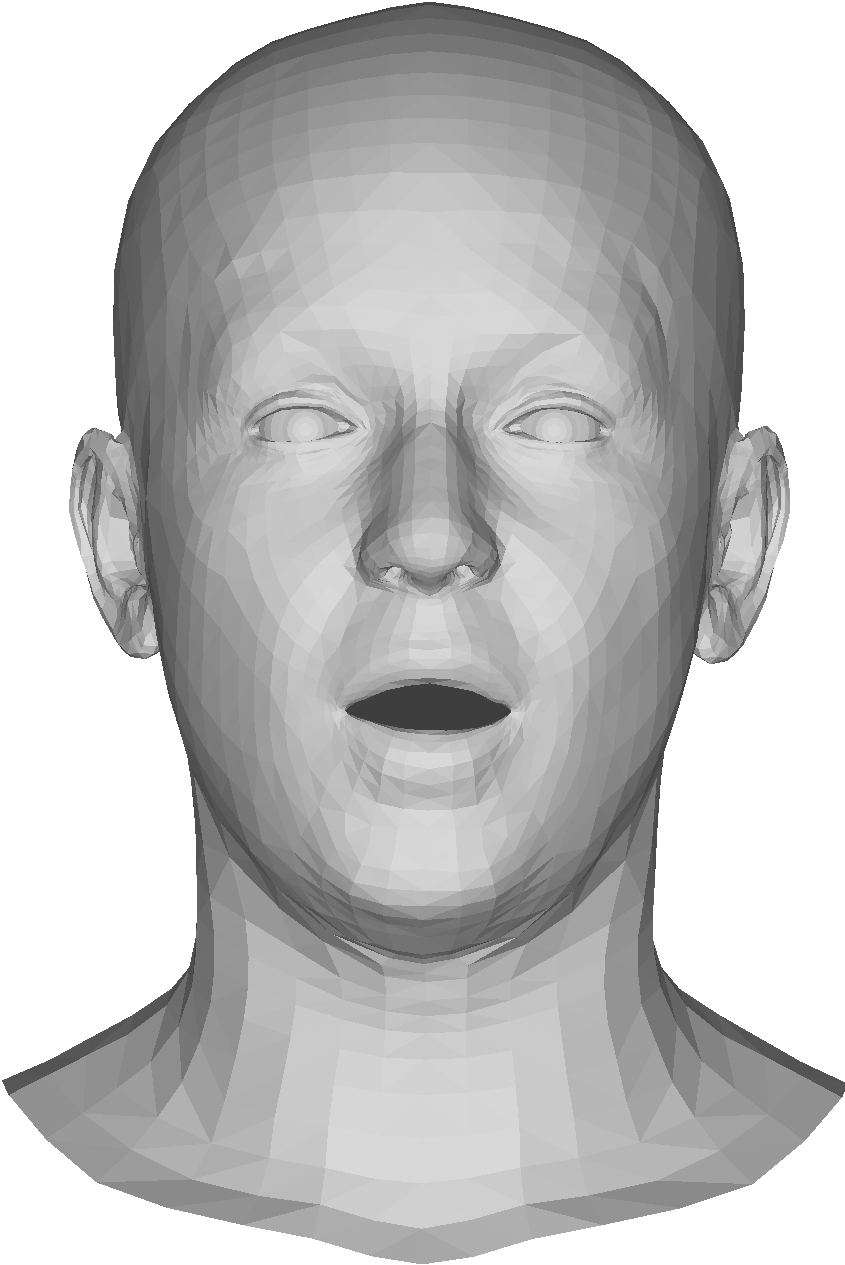} & \includegraphics[width=0.12\textwidth]{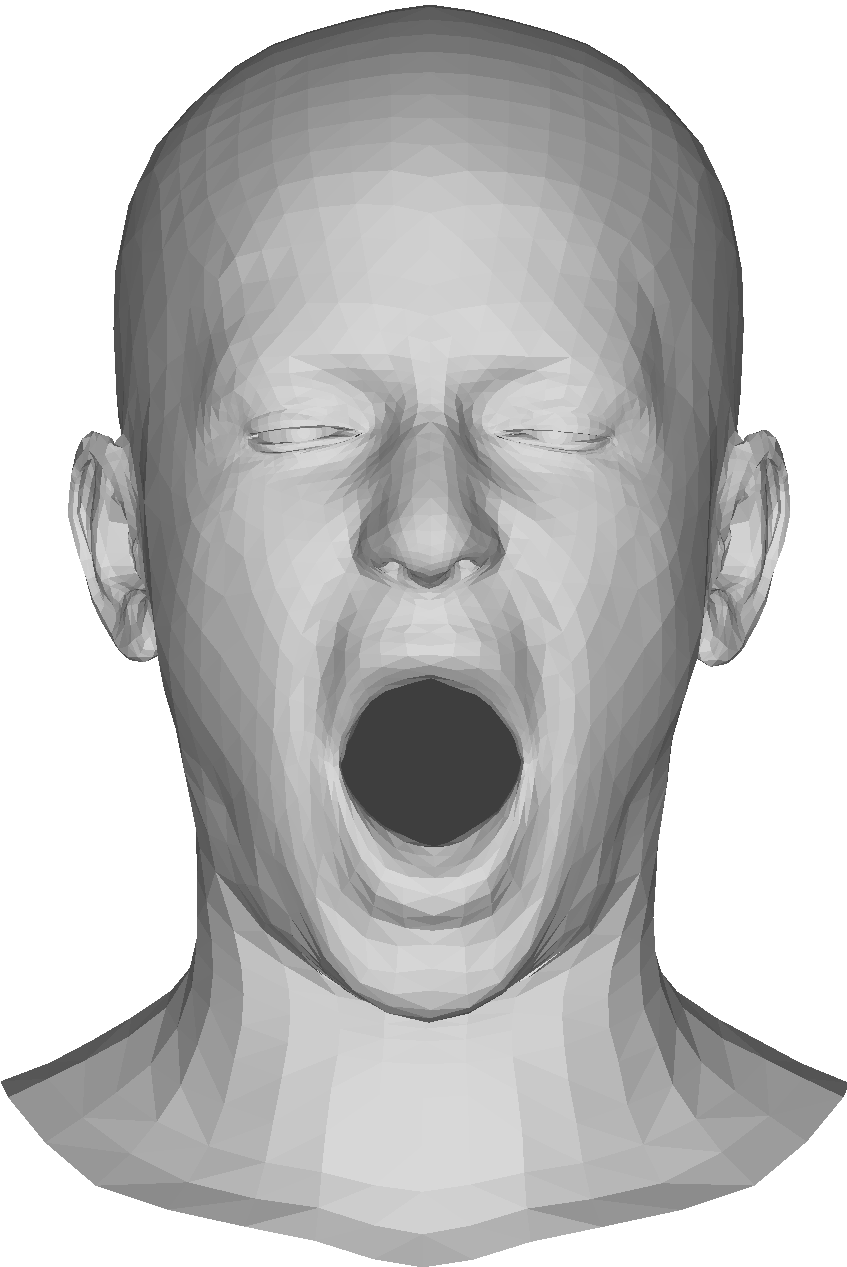} \\
    \\[-50pt] 
    &
    \includegraphics[width=0.12\textwidth]{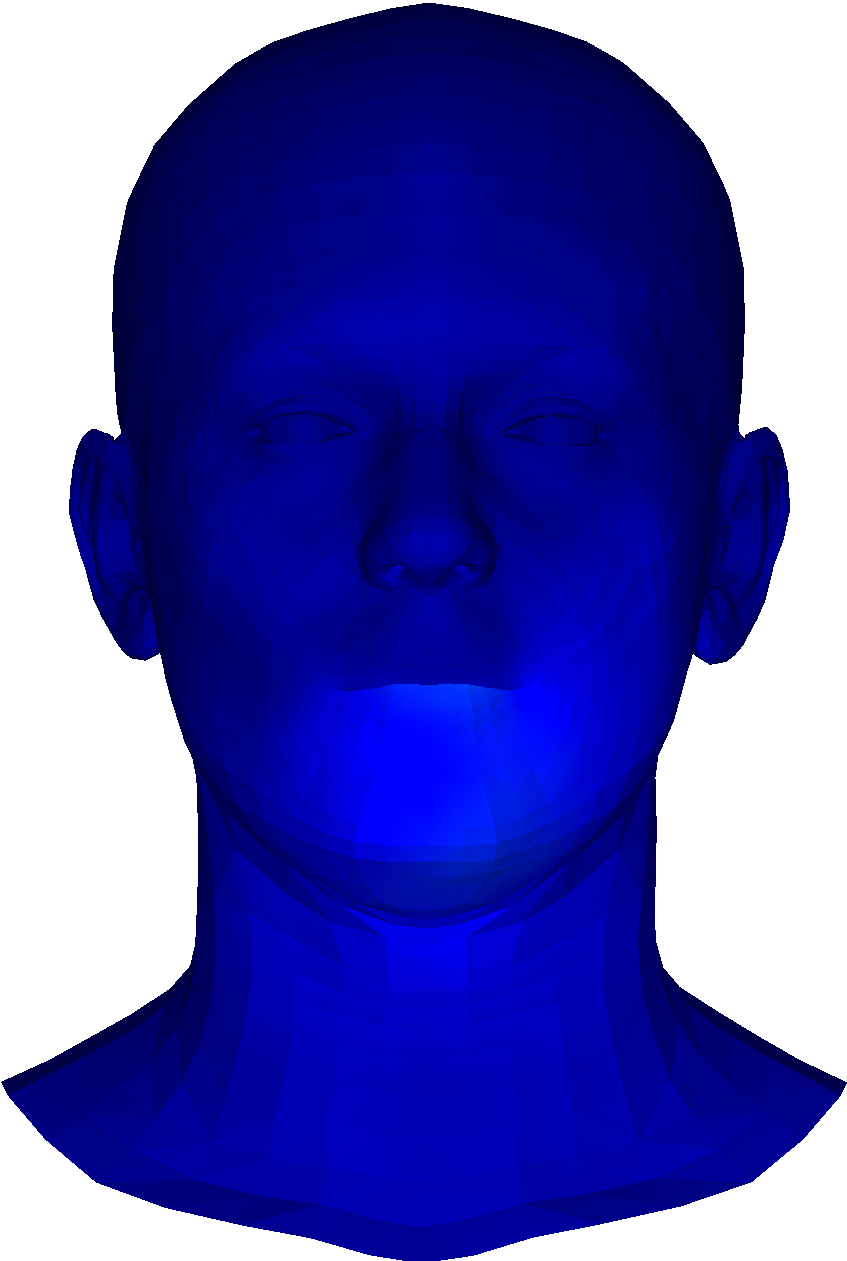} &
    \includegraphics[width=0.12\textwidth]{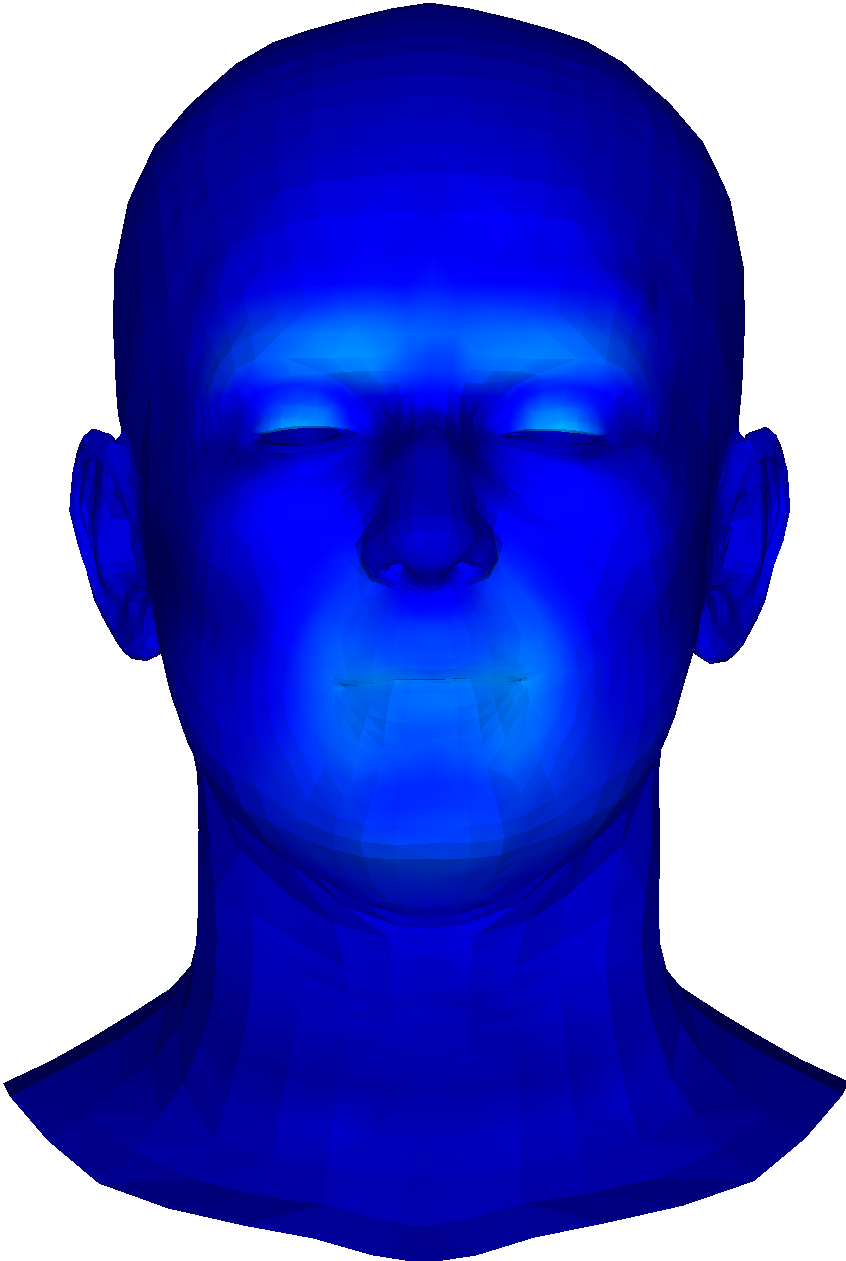} &
    \includegraphics[width=0.12\textwidth]{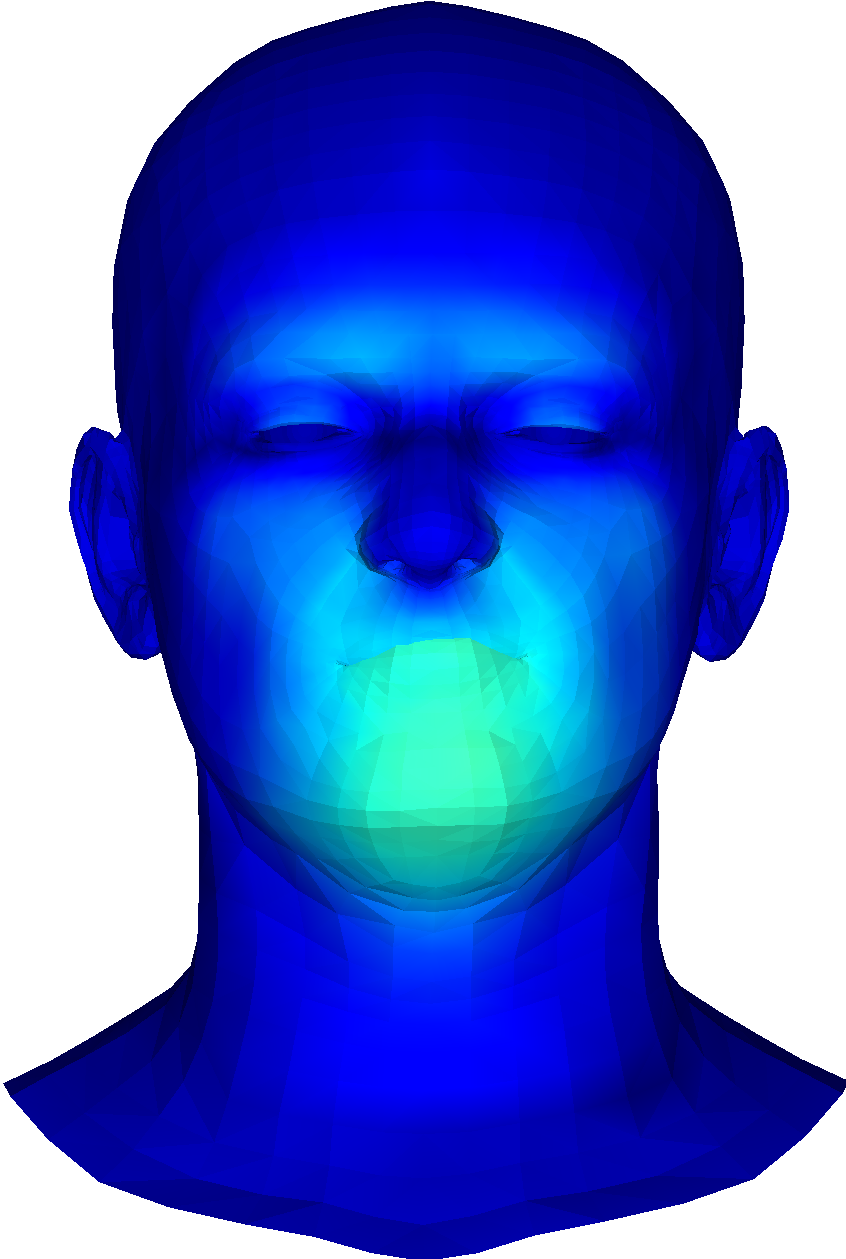} &
    \includegraphics[width=0.12\textwidth]{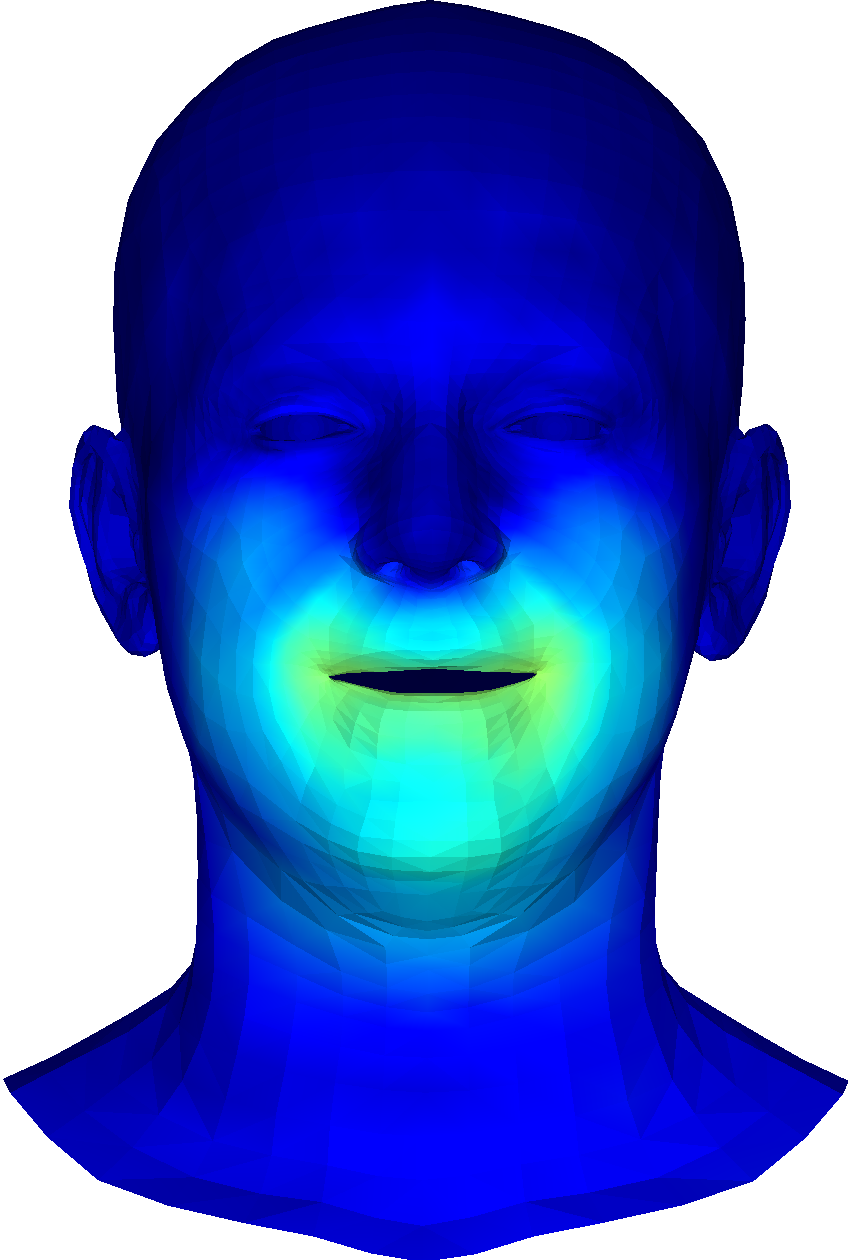} &
    \includegraphics[width=0.12\textwidth]{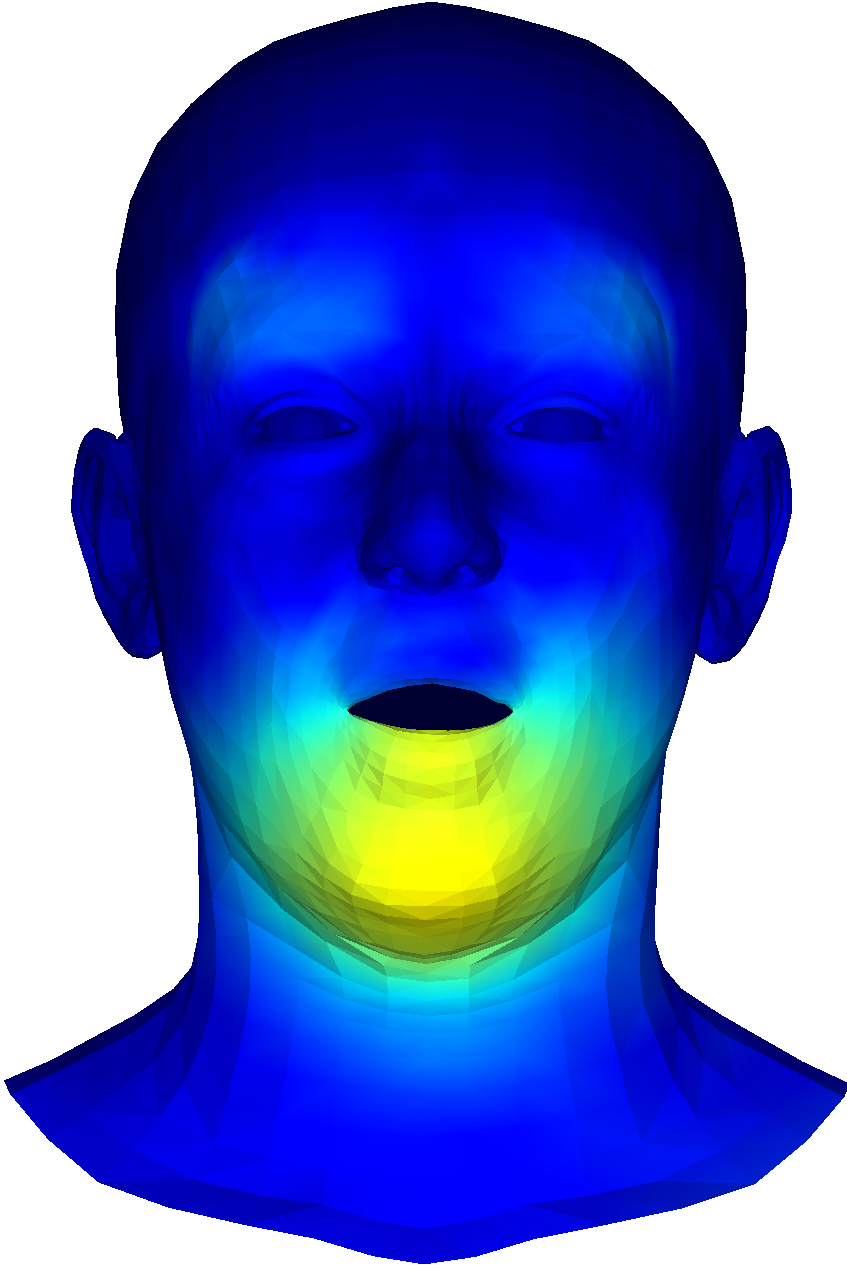} &
    \includegraphics[width=0.12\textwidth]{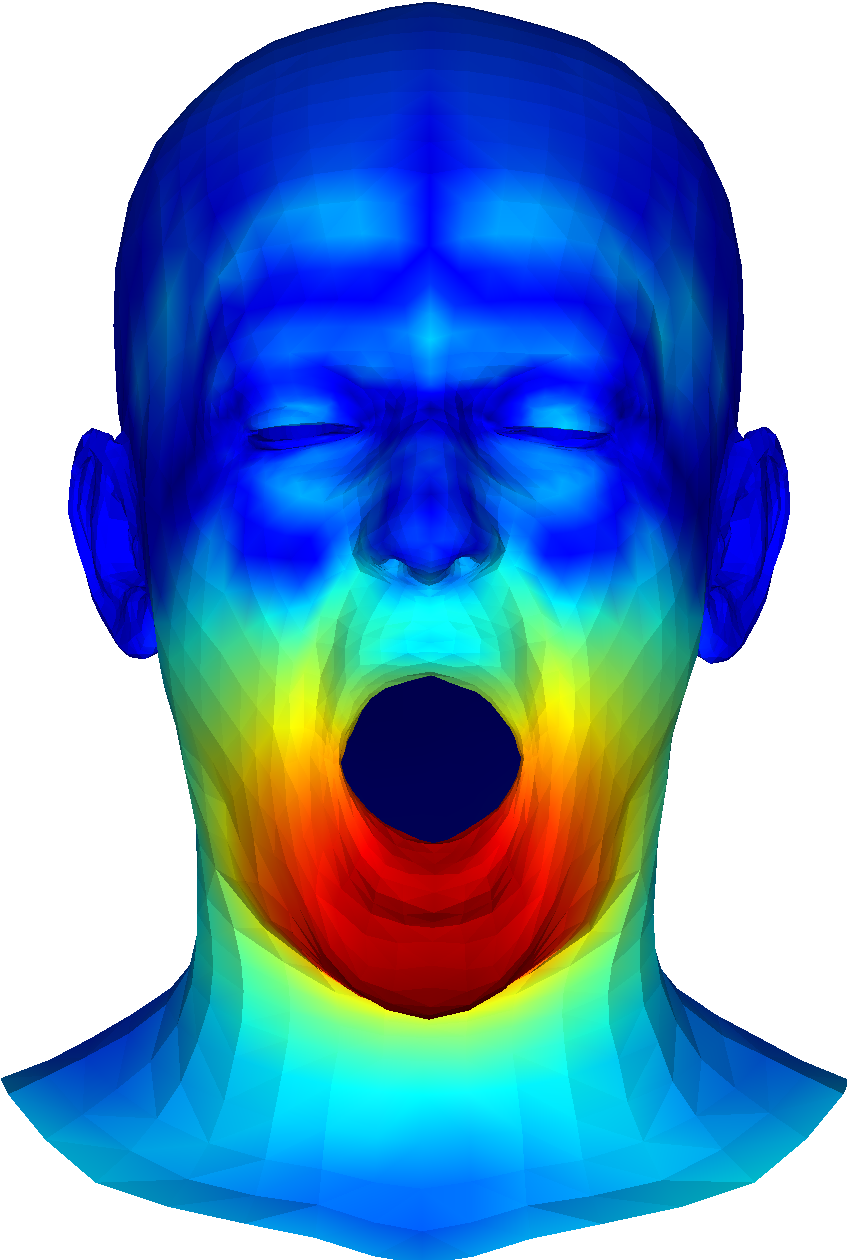}
    \\
  \end{tabular}
  \caption{Visualization of per-vertex Euclidean distances between the neutral anchor and the 3D face models of the average expression classes of the Multi-PIE dataset~\cite{Gross-MultiPIE-IVC-2010}.}
  \label{fig:class-avg-residuals}
\end{figure*}

Although aiming to predict face recognition utility, it is a necessary condition for our expression neutrality measures to carry relevant information for distinguishing neutral from non-neutral facial expressions. To validate the classification performance, Figure~\ref{fig:stacked-area-plots} depicts the detection error trade-off (DET) curves.

It is evident that the two-class SVMs achieve the best performance, with \textit{Detection Equal Error Rates (D-EERs)} of $9.1\%$ (Multi-PIE) and $10.7\%$ (FEAFA+). Generally, there is a discrepancy in the classification performance between the two evaluation datasets, which can be attributed to the more unconstrained capturing conditions of FEAFA+.

Note that classification performance is not indicative of the algorithms' performance in predicting face recognition utility. In classification, most errors occur when subtle facial expressions are falsely labelled as neutral.  However, subtle facial expressions have a limited impact on face recognition accuracy, as the deviation from expression neutrality is minimal. Consequently, it is crucial for the expression neutrality measures to identify expressions that deviate \textbf{significantly} from neutrality. These expressions are likely to have a more substantial impact on mated comparison scores.

We further present the class-related distributions of the neutrality quality measures on the Multi-PIE dataset in Figure~\ref{fig:stacked-area-plots}. A notable observation is that certain classes, such as \textit{scream} and \textit{surprised}, are dislocated further from expression neutrality than other classes by average. Additionally, it becomes clear that natural variation exists within each expression class, ranging from both subtle to more extreme instances.

\subsection{Explainable NeutrEx Measures}

\begin{figure*}
\centering
\begin{subfigure}{0.35\linewidth}
    \centering
    \includegraphics[width=\linewidth]{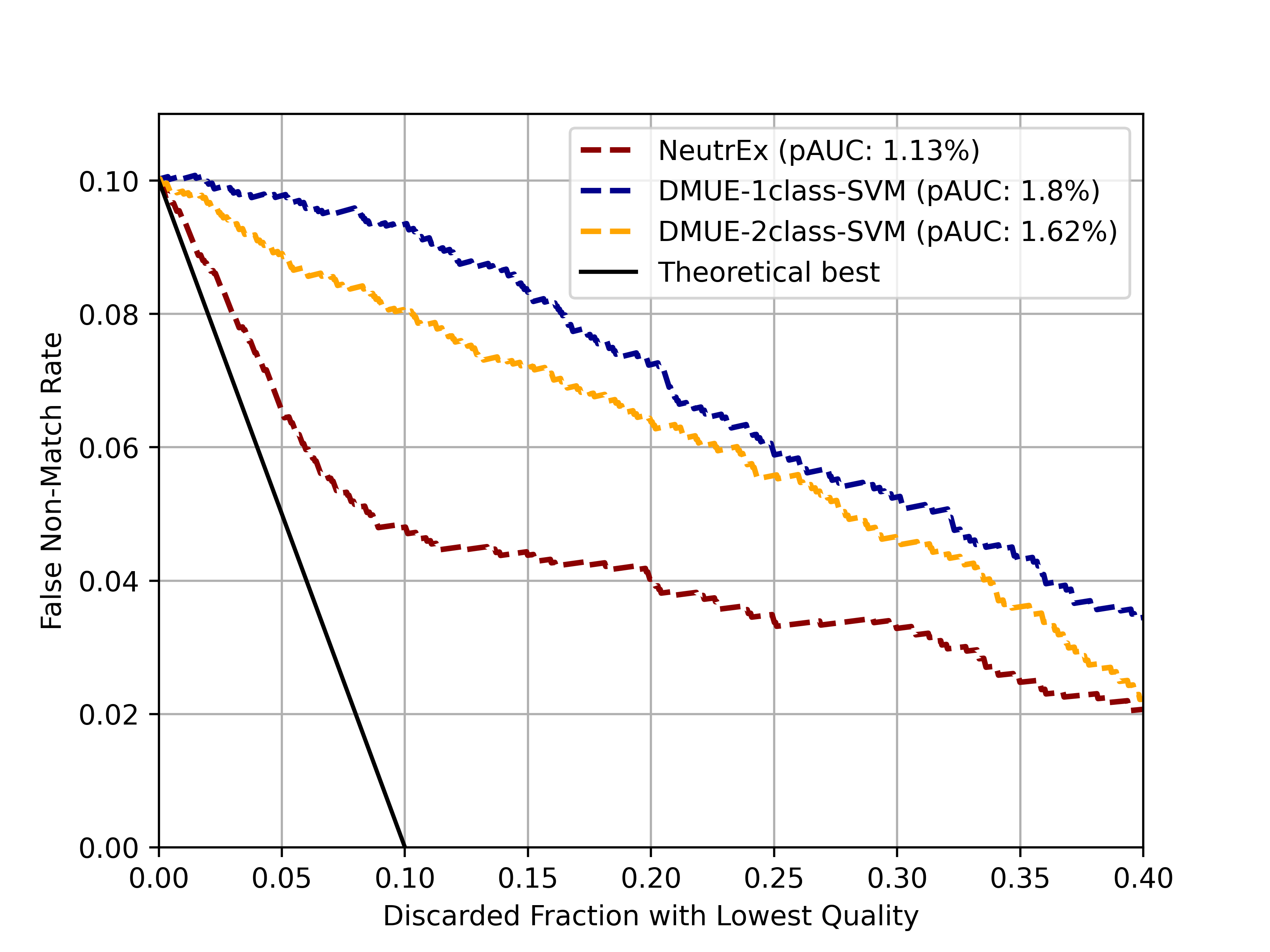}
\end{subfigure}
\begin{subfigure}{0.35\linewidth}
    \centering
    \includegraphics[width=\linewidth]{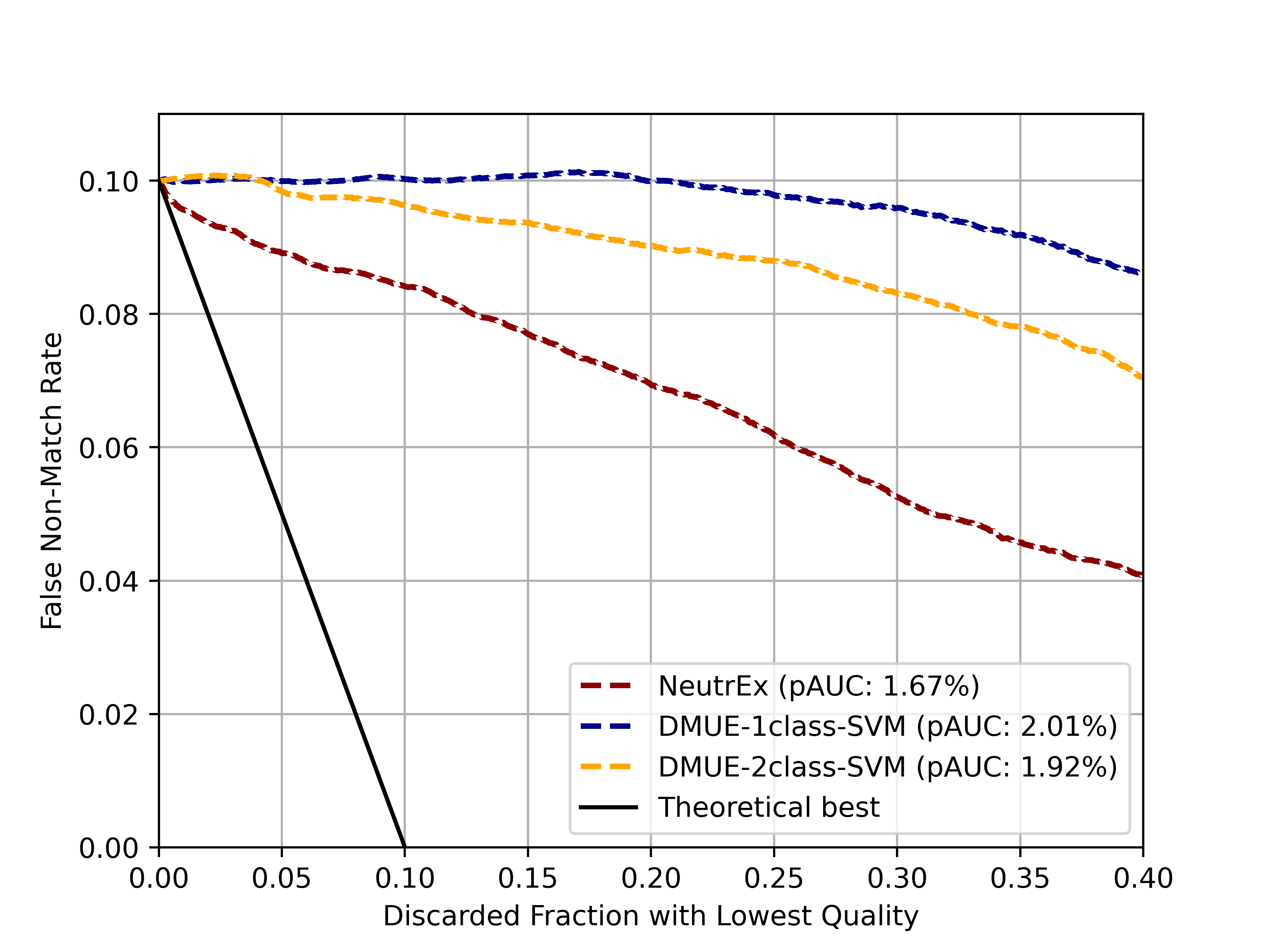}
\end{subfigure}

\begin{subfigure}{0.35\linewidth}
    \centering
    \includegraphics[width=\linewidth]{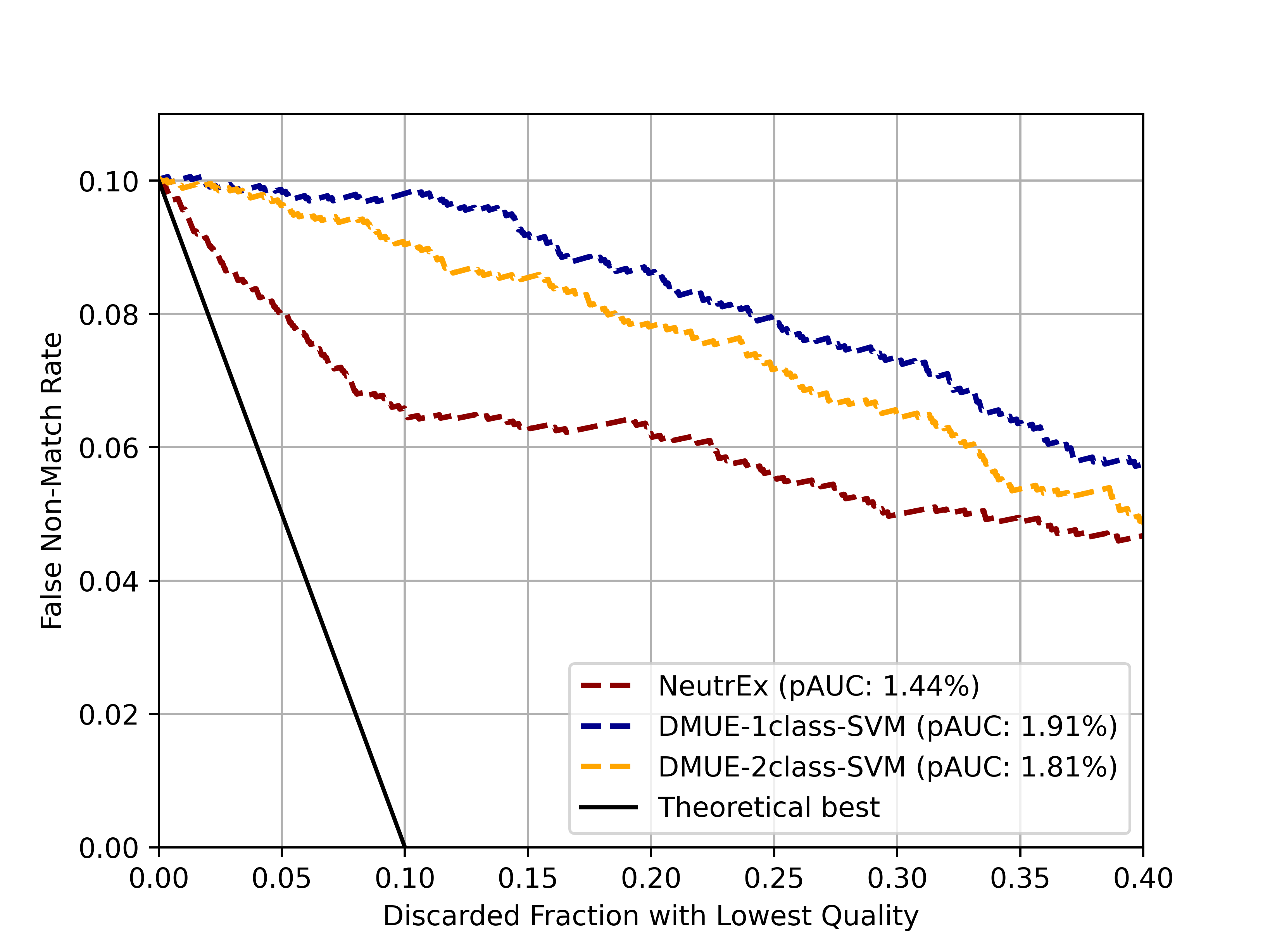}
    \caption{Multi-PIE}
\end{subfigure}
\begin{subfigure}{0.35\linewidth}
    \centering
    \includegraphics[width=\linewidth]{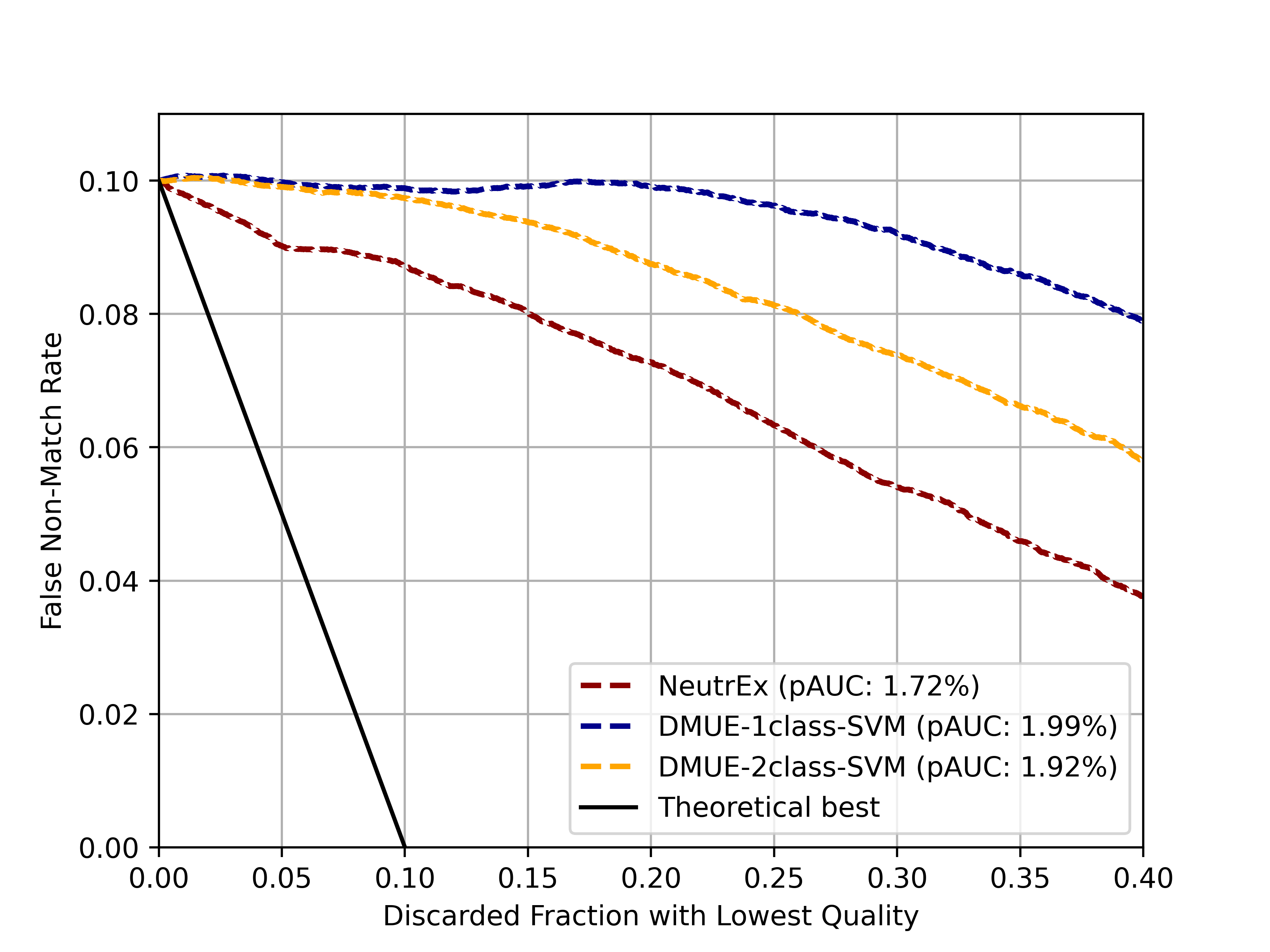}
    \caption{FEAFA+}
\end{subfigure}
\caption{Performance evaluation of our NeutrEx measure and baseline measures for predicting FR utility on MagFace~\cite{Meng-MagFace-CVPR-2021} (top) and Cognitec FaceVACS Version 9.6.0 (bottom) based on EDC curves. The pAUCs are reported in the range between 0\% and 30\% of discarded images estimated to be furthest apart from expression neutrality.}
\label{fig:edc-curves}
\end{figure*}

Complementary to the unified quality, component quality was introduced by the current committee draft of ISO/IEC 29794-5 by means of understanding the impact of an individual subject- or environmental-related attribute on the recognition outcome. Therefore, it is essential for expression neutrality measures to provide explainable predictions. This allows FR operators to provide actionable feedback to subjects automatically. 

In Figure~\ref{fig:stacked-area-plots}, the expression neutrality measures are visualized given ground-truth annotations, revealing which facial expressions typically cause large differences from neutrality. However, real-world applications require explaining the predictions of a single probe sample without manual labelling. By exploiting the full-vertex correspondence across the FLAME~\cite{Li-FLAME-ToG-2017} face models and the per-vertex distance computation of the NeutrEx measures, the deviation of individual face parts can be visualized to guide subjects when capturing their photos. In Figure~\ref{fig:class-avg-residuals}, the class-averaged 3D face models and their respective residuals to the neutral anchor are shown. Especially facial expressions with open mouths, such as \textit{surprised} and \textit{scream}, cause large residuals around the mouth and in the chin region, which is indicated by the high accumulated-vertex distance and low quality measure. As expected, the quality measure of the neutral class is ranked nearest to the neutral anchor derived from training data.

\subsection{Utility Prediction}
\label{sec:biometric quality assessment}

Figure~\ref{fig:edc-curves} shows the \textit{Error-vs-Discard Characteristic (EDC)} curves following the guidelines specified by ISO/IEC 29794-1~\cite{ISO-IEC-29794-1-QualityFramework-2023}. Generally, EDC curves visualize the performance of biometric quality assessment algorithms by systematically discarding images with the lowest estimated utility. The steepest curve with the lowest \textit{Partial Area Under Curve} (pAUC) represents the best algorithm for predicting the biometric performance in terms of the measured \textit{false non-match rate (FNMR)}. In this study, we restrict the pAUC calculation to a range of 0\% to 30\% to discard the lowest quality images while retaining those with medium or high quality. 

In Figure~\ref{fig:edc-curves}, we observe fundamental performance disparities in utility predictions on the Multi-PIE versus FEAFA+ datasets. Again, these differences can be attributed to the conditions under which the images were captured: The Multi-PIE dataset includes controlled variation in facial expression only. In contrast, FNMRs measured on FEAFA+ are also affected by other types of variation not reflected by our algorithms. This highlights the strong demand for adequate evaluation datasets, including controlled patterns relevant to the task.

Overall, the steep declines of the red curves in Figure~\ref{fig:edc-curves} indicate that facial expressions have a significant impact on the FR utility, particularly when deviating from expression neutrality. This finding validates the definition of expression neutrality as a quality component in ISO/IEC 29794-5.
 
Upon comparing the EDC curves, it becomes evident that our NeutrEx measures consistently outperform both baselines across all of the evaluated datasets and FR systems. When using MagFace, discarding $10\%$ of the Multi-PIE test samples reduces the FNMR to approximately $4\%$ based on our NeutrEx measures. In contrast, the two-class SVM requires the exclusion of $30\%$ of all samples to achieve an equal FNMR. This outperformance is also reflected by the lowest pAUC values of $1.13\%$ and $1.67\%$.

To explain the notable performance difference, we emphasize that DMUE~\cite{She-DMUE-CVPR-2021} was trained with the objective of optimizing classification performance quantified as class-wise confidence values. As such, both SVMs prioritize patterns relevant to facial expression classification but lack information about spatial differences. This means that images with subtle facial expressions might be sorted out by being labelled as non-neutral with high confidence, even though they are suitable for recognition. In summary, the comparison demonstrates that NeutrEx measures estimate utility more effectively and retain more data, which results in improved efficiency and a more convenient authentication mechanism.

\subsection{Limitations}
\label{sec:discussion}

While our NeutrEx measures significantly outperform the baselines, it is important to note the increased computational demands: The baseline approaches rely on a lightweight Resnet-18 network~\cite{He-Resnet-CVPR-2016} to compute the face embeddings, comprised of approximately $11$ million parameters. In contrast, adopting the pre-trained Resnet-50 networks~\cite{He-Resnet-CVPR-2016} of EMOCA.~\cite{Danvevcek-EMOCA-CVPR-2022} and DECA~\cite{Feng-DECA-ToG-2021} involves approximately $50$ million parameters in order to utilize the coarse shape and expression encoders. Nevertheless, we justify the allocation of more computational resources by emphasizing the substantial performance enhancement and improved explainability achieved through our proposed method.

\section{Conclusion}
\label{sec:conclusion}

In conclusion, the high effectiveness of our NeutrEx measure in serving as a utility predictor qualifies as a potential candidate algorithm for quantifying facial expression neutrality in the current working draft of \textit{ISO/IEC 29794-5}. With the transition from \textit{confidence-based} distances, as relied upon by our baseline approaches, to our NeutrEx \textit{per-vertex} distances, we estimate FR utility more precisely. In addition, we emphasize the high degree of explainability that enables FR practitioners to give actionable feedback to the subjects. A potential direction for future work might be to explore more suitable distance measures than using the Euclidean distance, focusing more on the \textit{topological differences} when comparing 3D face meshes. Further, future work could investigate the usage of teacher-student networks to reduce computational resources by training a more lightweight encoder-decoder architecture.

\section*{Acknowledgment}
This research work has been supported by the German Federal Ministry of Education and Research and the Hessian Ministry of Higher Education, Research, Science and the Arts within their joint support of the National Research Center for Applied Cybersecurity ATHENE.

{\small
\bibliographystyle{configs/ieee}
\bibliography{egbib}
}

\end{document}